\documentclass[authoryear, preprint,3p]{elsarticle} 

\usepackage{amssymb}
\usepackage{amsmath}
\usepackage{txfonts}
\usepackage{mathdots}
\usepackage{graphicx}
\usepackage{caption}
\usepackage{subcaption}
\usepackage{lineno,hyperref}
\usepackage{changepage}
\usepackage{setspace}
\usepackage{algorithm,algpseudocode}
\usepackage{enumerate}
\usepackage{tabularx,booktabs,caption,paralist}
\usepackage[table]{xcolor}
\usepackage{multirow,paralist} \usepackage{multirow}
\usepackage{float}
\usepackage{tabularx,booktabs,caption}
\usepackage{rotating}
\usepackage{appendix}

\DeclareGraphicsExtensions{.pdf,.png,.jpg}

\biboptions{authoryear}  

\begin{document}

\begin{frontmatter}



\title{Metaheuristic-based Energy-aware Image Compression for Mobile App Development}


\author[inst1]{Seyed Jalaleddin Mousavirad*}

\affiliation[inst1]{organization={Universidade da Beira Interior},
            city={Covilhã},
            country={Portugal},
        Email={, jalalmoosavirad@gmail.com}}

\author[inst2]{Luís A. Alexandre}
\affiliation[inst2]{organization={NOVA LINCS, Universidade da Beira Interior},
	city={Covilhã},
	country={Portugal},
	Email={, luis.alexandre@ubi.pt}}

\begin{abstract}
The JPEG standard is widely used in different image processing applications. One of the main components of the JPEG standard is the quantisation table (QT) since it plays a vital role in the image properties such as image quality and file size. In recent years, several efforts based on population-based metaheuristic (PBMH) algorithms have been performed to find the proper QT(s) for a specific image, although they do not take into consideration the user's opinion in advance. Take an android developer as an example, who prefers a small-size image, while the optimisation process results in a high-quality image, leading to a huge file size. Another pitfall of the current works is a lack of comprehensive coverage, meaning that the QT(s) can not provide all possible combinations of file size and quality. Therefore, this paper aims to propose three distinct contributions. First, to include the user's opinion in the compression process, the file size of the output image can be controlled by a user in advance. To this end, we propose a novel objective function for population-based JPEG image compression. Second, to tackle the lack of comprehensive coverage, we suggest a novel representation. Our proposed representation can not only provide more comprehensive coverage but also find the proper value for the quality factor for a specific image without any background knowledge. Both changes in representation and objective function are independent of the search strategies and can be used with any type of population-based metaheuristic (PBMH) algorithm. Therefore, as the third contribution, we also provide a comprehensive benchmark on 22 state-of-the-art and recently-introduced PBMH algorithms on our new formulation of JPEG image compression. Our extensive experiments on different benchmark images and in terms of different criteria show that our novel formulation for JPEG image compression can work effectively. Also, the experiments validate that the search strategy plays a crucial role in the performance of our new approach.   
\end{abstract}



\begin{keyword}
differential evolution\sep metaheuristic\sep particle swarm optimisation \sep  grey wolf optimiser \sep JPEG image compression.
\end{keyword}

\end{frontmatter}


\section{Introduction}
In recent years, a significant number of digital images are produced from cell phones, surveillance cameras, and personal digital devices. Popular apps such as Instagram and Tiktok, in particular, can handle the flow of millions of photos at once. These apps offer features related to image sharing. As a result, image compression is a crucial issue when it comes to storage space and network bandwidth usage. 

The most widely used method of compressing digital images is called JPEG (\textbf{J}oint \textbf{P}hotographic \textbf{E}xperts \textbf{G}roup), and it is based on the discrete cosine transform (DCT)~\cite{DCT_original}. There are several variants of JPEG such as the JPEG File Interchange Format (JFIF), as one of the most prevalent variants. Unlike JPEG standard, JFIF benefits from a color space. In other words, the first step in JFIF image compression is to convert the original color space to YCbCr color space, where Y, Cb, and Cr stand for luminance, blue, and red chrominance components, respectively. Each component is treated separately throughout the compression process. For simplicity, we will just employ the luminance component, Y; otherwise, the procedure is the same for all components. The Y component of the image is divided into $8 \times 8$ chunks, each of which is independently adjusted. The $8 \times 8$ blocks are zero-shifted before applying the DCT by deducting 128 from the element values. Each updated block is then quantised, leading to information loss. After quantisation, each block may be properly entropy encoded~\cite{JPEG01} without any data being lost. Depending on the quality factor value, different amounts of compression can be applied to an image.

A key component of the JPEG image compression is the quantisation table (QT). The luminance quantisation table (LQT) and chrominance quantisation table (CQT) are the two quantisation tables used by the Annex K variation~\cite{Annex_Jpeg}, the most significant variant of JPEG implementation. These two are primarily in charge of quantising the luminance and chrominance elements' respective DCT coefficient blocks. Since each image requires its own table, obtaining the right values for both quantisation tables is a tough and challenging task, as a result, most implementations utilise a typical value for the tables.

Metaheuristic algorithms (MA), and in particular population-based metaheuristic algorithms (PBMH), such as genetic algorithms (GA)~\cite{GA_Main_Ref} and particle swarm optimisation (PSO)~\cite{PSO_Main_Paper}, can be utilised to find the optimal QTs. PBMHs are iterative, stochastic, and problem-independent algorithms to direct the search process by using several operators toward an optimal point. A global optimum solution cannot be guaranteed by PBMHs, but they can offer a solution that is close to it~\cite{metaheuristics_talbi}.

One of the earliest attempts to use PBMHs for finding QTs~\cite{JPEG_GA} employed a GA algorithm to find the quantisation table such that the chromosome is an array of size 64 and the objective function is the mean square error between the original image and the compressed image. In another study, \cite{JPEG_GA2} used GA to create a JPEG image QT to compress iris images in iris identification systems. A knowledge-based GA was suggested by \cite{JPEG_GA3} to find the optimised quantisation table. To do this, the GA algorithm incorporates information regarding image properties and image compression. In another study~\cite{JPEG_DE01}, the quantisation table is designed using differential evolution (DE)~\cite{DE_Original}, and it is demonstrated that DE can outperform canonical GA. A knowledge-based DE is suggested by another study~\cite{JPEG_DE02} to enhance DE performance. In one of the most recent works~\cite{JPEG_KBS}, the rate-distortion optimal principle is taken into account for finding the QT(s), which offers a number of optimal solutions to the applications' need for multiple rates. Some other PBMHs which are used for finding QT(s) are the DE algorithm~\cite{JPEG_DE03}, the particle swarm optimisation (PSO)~\cite{JPEG_PSO}, the firework algorithm~\cite{JPEG_firework}, and the firefly algorithm~\cite{JPEG_FA01}.

Rate-distortion (RD) is a fundamental concept in image compression that involves trading off the amount of compression (rate) with the resulting image quality (distortion), and it is achieved through the quantisation step. While a variety of works focus in finding the RD-based image compression by using the conventional algorithms, there is not much work in PBMH-based image compression. \cite{JPEG_KBS} proposed a novel crossover and mutation based on the rate-distortion principle for the NSGA-II algorithm by changing the quantisation step, while \cite{JPEG_NSGA2_1} employed two conflicting objective functions, namely, compression rate and mean square error (MSE) for the multi-objective optimisation process based on NSGA-II algorithm. All works mentioned generate a set of solutions as the Pareto front. As a result, they are a member of a posteriori methods which the user can select the preferred solution among the generated solutions after the optimisation process. In other words, a user can not determine their preferences before conducting the optimisation process.

Even though images are one of the main energy consumers in smartphones, it might be difficult to estimate how much energy is used for any given process in a typical image. The majority of the present methods described in the literature can measure the power of a battery or, at best, for a particular application~\cite{Petra}. To tackle this issue, \cite{JPEG_Postdoc01} employed the energy profiler of Android Studio and Plot Digitiser software to show that image quality and file size have a vital influence in the energy usage of an application. Therefore, they introduced the concept of "Energy-aware JPEG Image Compression" in the sense of methods that can save energy consumption in a device such as a smartphone. To this end, they showed that file size and image quality are two proxies for the energy consumption in Android smartphones. Developers frequently strive for both reduced file sizes and better image quality, however these two goals are incompatible since better image quality will increase file size and energy usage. Hence, a compromise between image quality and file size is required. To tackle this issue, \cite{JPEG_Postdoc01} proposed two general multi-objective approaches: scalarisation and Pareto-based. They employed five scalarisation algorithms, including GA, PSO, DE, evolution strategy (ES), and pattern search, while two Pareto-based methods, including Non-Dominated Sorting Genetic Algorithm II (NSGA-II) and a reference-point-based NSGA-II (NSGA-III) are used for the embedding scheme.   

The above-mentioned studies have focused on finding an optimal QT, while they suffer from a few fundamental problems that, to the best of our knowledge, no PBMH-based research has yet addressed, including 1) ignoring the user's opinion in advance, 2) lack of comprehensive coverage, and 3) lack of sufficient knowledge of a user to determine the quality factor.

The first problem with the current studies is that they ignore the user's opinion in advance. Different PBMH algorithms try to find the proper QT(s) based on one (or more) criterion, while the output image may not have the features required by the user. Assume that an Android developer is considering a small-size app. The optimisation algorithms for this app endeavor to produce a high-quality image that results in a larger file size. As a result, the production of such an image might be contrary to the opinion of the user, who prefers an image with a small file size. Although there are few research that consider the user's opinion, it should be noted that these methods first perform the optimisation process (a posteriori optimisation) and then provide a  set of solutions. The user must choose one of these solutions, while the generated solutions sometimes is sparse~\cite{JPEG_Postdoc01}. In addition, the scalarisation methods in the literature tries to decrease the file size regardless of what size a developer needs. Therefore, developing a user-specified file size for PBMH-based JPEG image compression is indispensable, so that a user can determine the file size for a specific image in advance, before starting the optimisation process (a priori optimisation). The second problem with the population-based JPEG image compression is the lack of comprehensive coverage. Our experiments (Section~\ref{sec:QF}) imply that working on only the QT cannot provide all possible combinations of file size and quality. This issue becomes acute when the goal is to find a specific combination of file size and image quality that cannot be achieved by only using QTs. The third problem comes from a lack of enough knowledge of a user to determine the quality factor. To the best of our knowledge, these three problems have not yet been studied in population-based JPEG image compression. 

Therefore, the main contributions of this paper is as follows:
\begin{itemize}
	\item The first contribution incorporates a user-specified file size to include the user's opinion in our proposed approach. To this end, we propose an objective function for metaheuristic algorithms that first tries to produce an output image as close to the user-specified file size, and secondly has the highest image quality at the same time. 
	\item  To provide comprehensive coverage, we proposed a novel representation for PBMH-based image compression by adding only one component to enhance the amount of coverage. 
	\item The proposed algorithm has the capability to find the quality factor automatically without any reference image and prior knowledge about the color distribution of an image. 
	\item We perform a comprehensive evaluation of PBMH-based search strategies for our re-formulated JPEG image compression. For this purpose, we select a total of 22 algorithms, including base, advanced, and metaphor-based algorithms. Some of the selected algorithms are state-of-the-art such as DE and its variants, while some others are among the more recent algorithms that have already received significant attention (based on paper citations). 
\end{itemize}

The main characteristics of our approach are summarised below. 
\begin{itemize}
	\item Our novel representation is simple but effective since it can give comprehensive coverage in the search space and its implementation is straightforward.
	\item Our proposed objective function is simple to implement.
	\item The approach is independent of the PBMHs since it is based on representation and objective function and not a search strategy. As a result, it works with all PBMHs. 
	\item The three disadvantages of population-based JPEG image compression are solved without any further function evaluation. As a result, we can say that the time complexity remains approximately the same.
	\item Our proposed algorithm focuses on different aspects of JPEG image compression in a single shot, meaning that it includes user-specified file size, high-quality image, capability to find the right values for QTs,  and the ability to find the quality factor automatically.
\end{itemize}

The remainder of the paper is organised as follows. Section~\ref{sec:JPEG} briefly describes the JPEG Image Compression. Section~\ref{sec:proposed} defines our novel representation, novel objective functions, and the search strategies for re-formulated JPEG image compression, while Section~\ref{sec:Exp} discusses the results. Finally, Section~\ref{sec:Conc} concludes the paper.

\section{The JPEG Image Compression}
\label{sec:JPEG}
\begin{figure}[tb]
	\centering
	\includegraphics[width=.9\columnwidth]{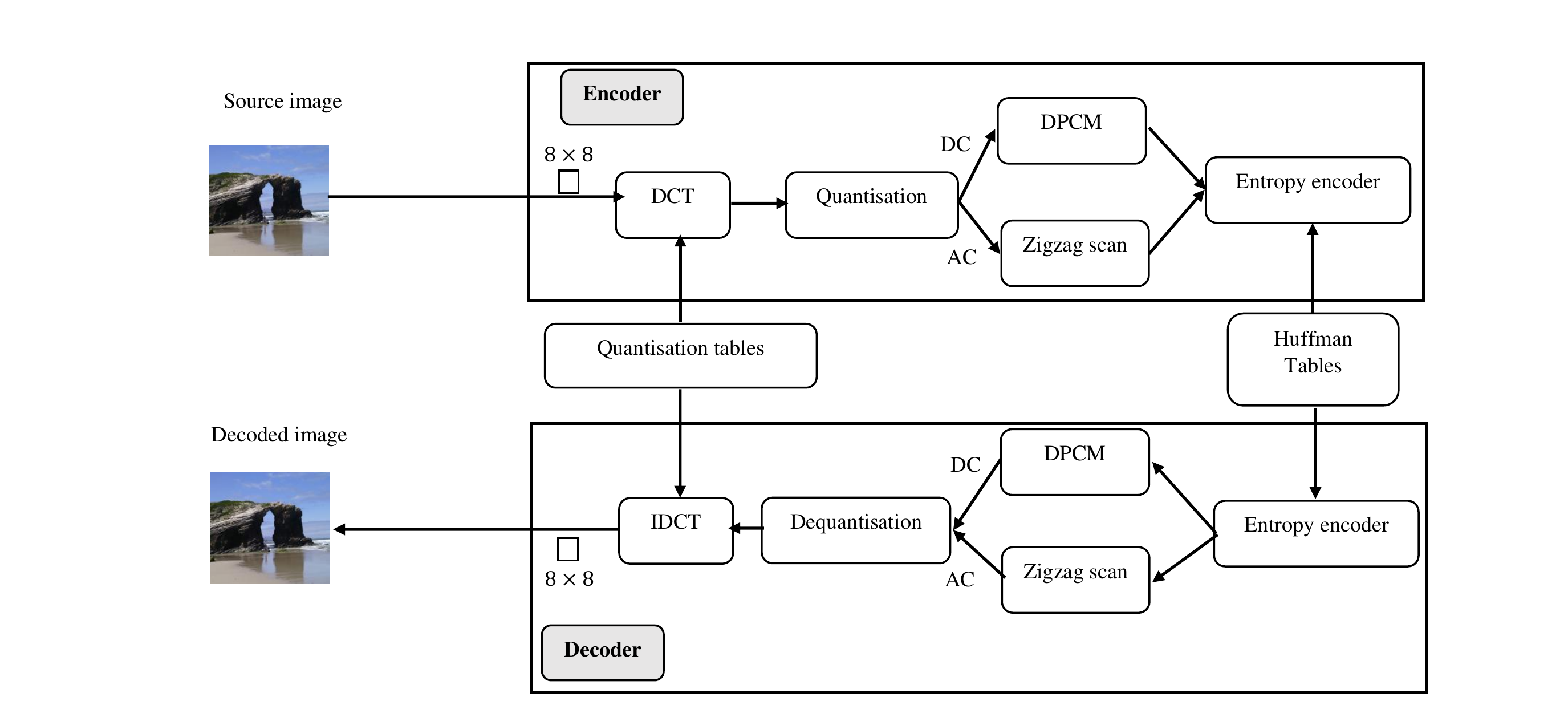}
	\caption{The general structure of JPEG Image Compression.} 
	\label{fig:JPEG}
\end{figure}

The essential elements of JPEG image compression are shown in Figure~\ref{fig:JPEG}. The encoder is in charge of transforming the original image into the JPEG compression variation, while the decoder is in charge of doing the opposite. A JPEG file can be encoded in a variety of ways, but JFIF (JPEG File Interchange Format) encoding is one of the most prevalent. The first step in JFIF is the color representation which is changed from RGB to YCbCr, which consists of one luma component (Y), which stands for luminance, and two chroma components (CB and CR), which stand for blue and red chrominance components. The other parts of JFIF are almost similar to the standard JPEG. We go into further depth on the key elements below.

Blocks of $ 8 \times 8$ pixels are initially created from the original image. After that, the block values are moved from $[0,2p-1] $ to $[-2^{p-1},2^{p-1}-1] $, where $p$ is the number of bits per pixel ($p=8$ for the standard JPEG compression). A vector with a size of $64 \times 1$ should be provided into the Discrete Cosine Transform (DCT)~\cite{DCT_Journal} component for each block of $8 \times 8$ pixels. The DCT block divides the input signal into 64 DCT coefficients, or basis-signal amplitudes. The DCT can be mathematically written as 
\begin{equation}
\label{eq:DCT}
F(u,v)=\frac{1}{4}c_{u}c_{v}  \left[ \sum_{x=0}^{7} \sum_{y=0}^{7} f(x,y) \cos \left( \frac{(2x+1)u\pi}{16} \right) \cos \left(\frac{(2y+1)v\pi}{16} \right) \right]
\end{equation}
where
\begin{equation}
c_{r}=\begin{cases}
\frac{1}{\sqrt{2}} & r=0 \\
1 & r>0
\end{cases} 
\end{equation}
The reversal of the DCT component used to rebuild the original image is called Inverse DCT (IDCT), and it is described as
\begin{equation}
\label{eq:DCT}
F(x,y)=\frac{1}{4}c_{u}c_{v} \left[ \sum_{u=0}^{7} \sum_{v=0}^{7} f(u,v)  \cos  \left(\frac{(2x+1)u\pi}{16} \right) \cos \left(\frac{(2y+1)v\pi}{16} \right) \right]
\end{equation}

The original 64-point signal is exactly recovered in the absence of the quantisation step.

\subsection{The Quantisation and Dequantisation Components}

The 64-element quantisation table, which is used in the quantisation stage, should be known beforehand. Each item in the table that corresponds to [1,255], the baseline, specifies the step size of the quantiser for its associated DCT coefficient. By deleting information that is not visually significant, quantisation seeks to accomplish compression while retaining picture quality.

The definition of the quantisation image is
\begin{equation}
L(u,v)= round\left(  \frac{F(u,v)}{Q(u,v)}\right)
\end{equation}

where $Q(u,v)$ denotes the corresponding entry of the quantisation table, $L(u,v)$ stands for the quantised DCT coefficients, $F(u,v)$ stands for the DCT coefficients, and $round(x)$ is the closest integer number to $x$. It is important to note that the information loss increases with the value of $Q(u,v)$.

A preliminary approximation of $F(u,v)$ is recreated by the de-quantisation component of the decoder by reversing the quantisation process as
\begin{equation}
\bar{F}(u,v)=L(u,v) \times Q(u,v)
\end{equation}

Since the quantisation table causes an information loss, this step is essential to the JPEG compression process. In order to balance the quality of the reconstructed image with the efficacy of the compression, the quantisation table plays a crucial role in JPEG image quantisation.

The quality factor can be added as a component of the JPEG implementation~\cite{JPEG_NSGA2_1}. For a given quality factor $F$, the elements of the associated QT are obtained as
\begin{equation}
Q_{i,j}=\left[\frac{50+S+Q_{i,j}}{100}\right]
\end{equation}
where $[x]$ means the rounding of $x$, $Q_{i,j}$ is the QT in the location of $(i,j)$, and $S$ is defined as
\begin{equation}
S=\begin{cases}
200-2F & F\geq 50 \\
\frac{5000}{Q} & else
\end{cases} 
\end{equation} 
\subsection{Symbol Coding }
After quantisation, the $8 \times 8$ block's 63 AC coefficients are treated separately from the DC coefficient. The DC coefficient is encoded using the Differential Pulse Code Modulation (DPCM) as

\begin{equation}
DIFF_{i}=DC_{i}-DC_{i-1}
\end{equation}
where $DC_{i}$ and $DC_{i-1}$ are the DC coefficients for the current $8 \times 8$ block and the preceding $8 \times 8$ block, respectively.

The quantised 63 AC coefficients may be formatted for entropy coding using a zigzag scan~\cite{JPEG_Zigzag}. The AC coefficients after the zigzag scan exhibit decreasing variances and increasing spatial frequencies.

\subsection{Entropy Coding }
There are frequently a few nonzero and a few zero-valued DCT coefficients left behind after the quantisation procedure. Entropy coding aims to compress the quantised DCT coefficients by their statistical characteristics. JPEG uses Huffman coding as its default method, which uses two DC and two AC Huffman tables for the luminance and chrominance DCT coefficients, respectively~\cite{JPEG_Zigzag}.

\section{Re-formulated Population-based JPEG Image Compression}
\label{sec:proposed}
This paper proposes a re-formulated population-based JPEG image compression. To this end, we propose, for the first time, a population-based strategy to find an JPEG image so that the size of the output file is as close as the user-specified file size, but with the highest image quality. Also, we propose a novel representation so that the search space could be covered better and the quality factor could be found automatically. Generally speaking, a representation, an objective function, and a search strategy  are the three primary considerations when using a PBMH method to solve an optimisation problem. Representation exhibits the structure of each candidate solution, while an objective function is responsible for quantifying the quality of a candidate solution. The search strategy aims to find a promising solutions using several operators. This paper mainly proposes a novel representation and also a novel objective function with the aim of finding an image with a file size as close as to the user-specified file size, while maintaining the highest image quality. In addition, we selected 22 algorithms, which not only allows us to choose the best strategy, but also provided a benchmark among different algorithms.

In this section, first, we conduct a behaviour analysis of the quality factor, and then explain the main components of the proposed strategy.
\subsection{Behaviour Analysis of the Quality Factor}
\label{sec:QF}
In this section, we conducted a behaviour analysis on the quality factor (QF). For all experiments, we randomly generated 10000 QTs and QFs to build a JPEG image and then calculated PSNR and the file size for the output image. 

\begin{figure}[tb]
	\centering
	\begin{subfigure}[b]{0.49\linewidth}
		\includegraphics[width=\linewidth] {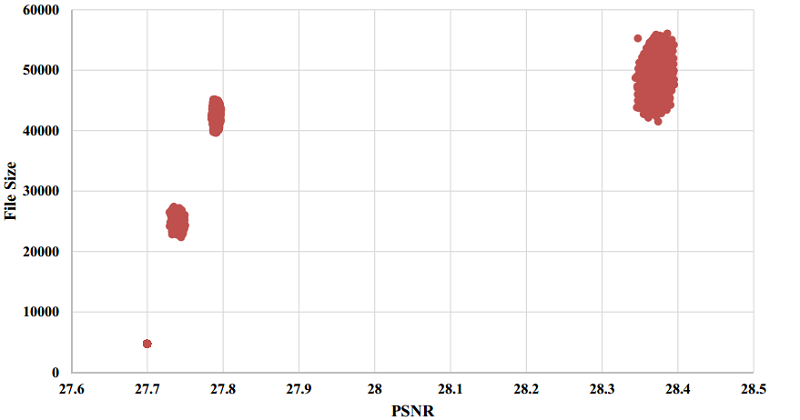}
		\caption{}
	\end{subfigure} 
	\begin{subfigure}[b]{0.49\linewidth}
		\includegraphics[width=\linewidth] {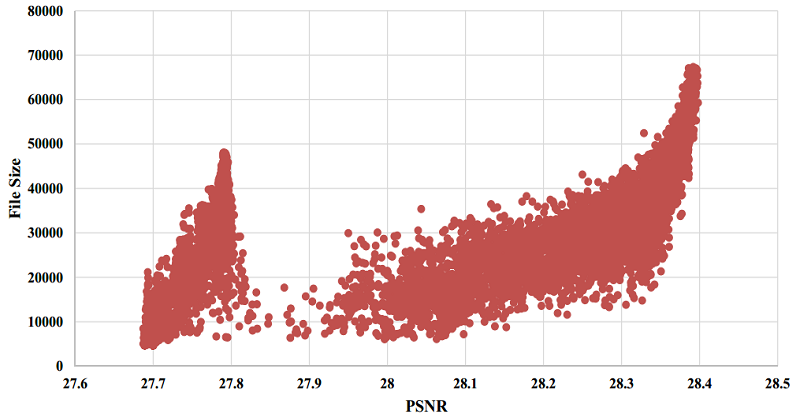}
		\caption{}
	\end{subfigure}
	\begin{subfigure}[b]{0.49\linewidth}
		\includegraphics[width=\linewidth] {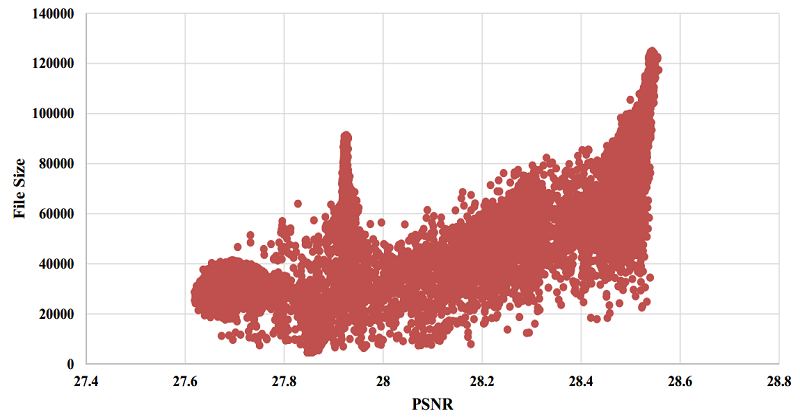}
		\caption{}
	\end{subfigure} 
	\caption{(a): Data Distribution in terms of PSNR and file size over random numbers for QTs with permutation; (b) Data Distribution in terms of PSNR and file size over random numbers for QTs without permutation; (c) Data Distribution in terms of PSNR and file size over random numbers for QTs without permutation but including quality factor.  }
	\label{fig:qf}
\end{figure}

In the first experiment, we randomly generated the QTs 10000 times using permutation. In other words, the values for each QT are unique, and there are no duplicates. For a typical image, the data distribution in terms of PSNR and file size can be observed in Figure~\ref{fig:qf}.a . It is clear that, in this case, the search space is divided into several clusters, and the generated data does not cover the search space entirely.

In the next experiment, we added QF to the experiments, meaning that QF was not a fixed number and selected as a random integer. Figure~\ref{fig:qf}.b shows that by adding QF, the search space has more coverage.

In the third experiment, we selected QFs without permutation, meaning that the values for QTs can be duplicate. From Figure~\ref{fig:qf}.c, the result is impressing, indicating that the search space is comprehensively covered. Therefore, the two leading factors that are very effective in covering the search space are QF and not using permutation. In our proposed algorithm, these two are embedded in our representation strategy.
\subsection{Solution Representation}
The conventional representation for population-based JPEG image compression is a vector as
\begin{equation}
x=[{QT_{1},QT_{2},...,QT_{m}}]
\end{equation} 
whose length is the number of elements in the QT(s) (it is 64 per table) and $Qt_{i}$ indicates the $i$-th element. This representation is so prevalent in the literature, and to the best of our knowledge, most research uses this representation. In this representation, only QT(s) are encoded and the goal of optimisation is to identify the best elements for QT(s) .

This paper, first, proposes a novel representation that encodes not only the QT(s) but also the quality factor. The representation is a vector of integer numbers of dimension 129 as
\begin{equation}
x=[LQT_{1,1},...,LQT_{8,8},...,CQT_{1,1},...CQT_{8,8}, QF]
\end{equation}   
where the $LQT_{i,j}$ and $CQT_{i,j}$ denote the element that belongs at the coordinates $(i,j)$ in the LQT matrix and the CQT matrices, respectively. In other words, the LQT table's first 64 entries are positive integer integers in the range $[0,2p-1]$ (where $p$ is the number of bits corresponding to a pixel; in our case, $p=8$), while the remaining elements are set aside for the CQT table. The last entry is an integer number in $[1,99]$, which controls the quality factor. Our new representation is able to find the QF automatically since it is a part of the candidate solution. Also, by adding QF to the representation, the search space can be covered entirely. 

It is worth noting that the conventional search space has a size of $256^{128}$, whereas the new search space has a size of $256^{128} \times 99$. In other words, even though we only added one variable to our new representation, the search space has grown by 99 times. 
\subsection{Objective Function}
One of the main goals of this paper is to present a user-specified population-based JPEG image compression. To this end, the desired file size is recommended by the user. In other words, there are two specific purposes that must be pursued in the objective function, as follows:
\begin{enumerate}
	\item the output file size of image should be as close as possible to the file size specified by the user.
	\item the image quality should be maximised.
\end{enumerate} 

To achieve the first purpose, the following function can be defined, which is the distance between the output file size and the user-specified file size

\begin{equation}
\label{eq:obj1}
obj = \frac{|FS_{US}-FS_{O}|}{FS_{US}}
\end{equation}

where $FS_{US}$ shows the user-specified file size, while $FS_{O}$ is the file size of output image. Also, for normalisation, this difference is divided by $FS_{US}$.

The second purpose can be achieved by maximising a quality factor such as PSNR. Therefore, the final objective function can be defined as

\begin{equation}
\label{eq:obj}
obj = \frac{|FS_{US}-FS_{A}|}{FS_{US}}+\frac{\lambda}{PSNR}
\end{equation}
where $\lambda$ is a parameter, aiming to normalise the two goals in an almost identical range.
\subsection{Search Strategies}

For search strategies, we can use any type of PBMHs. It is obvious that we cannot analyse every PBMH technique that has been published in the literature due to the large and diverse collection. Therefore, based on two criteria, we chose a number of algorithms for our investigation. While certain algorithms, like GA and DE, are state-of-the-art techniques that are often used in evolutionary and swarm computing, others, like the grey wolf optimiser (GWO)~\cite{GWO_Main_Paper}, are more recent algorithms that have still drawn substantial attention based on paper citations. 

Eventually, we selected 22 algorithms, classified into three different categorises: base algorithms, advanced algorithms, and metaphor-based algorithms. In the following, we briefly explain the algorithms, while we refer to the cited publications for more details.

\subsubsection{Base Algorithms}
\begin{compactitem}
	\item
	\textbf{Genetic algorithm (GA)~\cite{GA_Main_Ref}:} GA is the oldest population-based algorithm and has two main operators, crossover and mutation. Crossover combines the information from the parents, while mutation makes random changes to one or more elements of a candidate solution. Solutions are carried over from one iteration to the next based on the principle of ``survival of the fittest''. GA uses selection operators, both for choosing the parents for crossover and mutation and for choosing the solutions that pass to the next generation
	
	\item
	\textbf{Differential Evolution (DE)~\cite{DE_Original}:} DE has three main operators, mutation, crossover, and selection. Mutation generates candidate solutions based on a scaled difference among candidate solutions and generates a mutant vector, \textit{DE/rand/1},  as
	\begin{equation}
	v_{i}=x_{r1} + F (x_{r2}-x_{r3}) ,
	\end{equation} 
	where $F$ is a scaling factor, and $x_{r1}$, $x_{r2}$, and $x_{r3}$ are three different randomly selected candidate solutions from the current population. Crossover integrates the mutant vector with a target vector selected from the current population. Finally, a candidate solution is selected by a selection operator.
	\item \textbf{Memetic Algorithm (MA)~\cite{Memetic_Main}}: MA is a population-based search strategy that uses a population-based algorithm (here GA) in the combination with a local search. In the version we used, there is a probability for each agent, indicating whether a local search should be done or not.  
	\item
	\textbf{Particle Swarm Optimisation (PSO)~\cite{PSO_Main_Paper}}: it is a swarm-based optimisation technique whose updating process is based on the best position of each candidate solution and a global best position. The velocity  vector of a particle is updated as   
	\begin{equation}
	\label{Eq:PSO}
	v_{t+1}= \omega v_{t}+c_{1} r_{1} (p_{t}-x_{t})+c_{2} r_{2} (g_{t}-x_{t}) ,
	\end{equation} 
	where $t$ is the current iteration, $r_{1}$ and $r_{2}$ are random numbers from a uniform distribution in $[0;1]$, $p_{t}$ is the personal best position, and $g_{t}$ is the global best position. 
	\item 
	\textbf{Evolutionary strategy (ES)~\cite{ES_main_paper}:} ES is a metaheuristic algorithm where each offspring is generated based on a Gaussian random number as
	\begin{equation}
	\label{Eq:ES}
	x_{new}= x_{old}+N(0,\sigma^{2}) ,
	\end{equation} 
	where $N(0,\sigma^{2})$ is a  Gaussian random number with mean 0 and variance $\sigma^{2}$. Then, competition should be done for each individual and finally, the best individuals transfer to the next generation.
	\item
	\textbf{Artificial Bee Colony (ABC)~\cite{ABC_Main_Paper}}: it mimics the foraging behaviour of honey bees. There are three types of bees, employed bees, onlookers, and scouts. Each employed bee, $i$, generates a candidate solution as    
	\begin{equation}
	\label{eq:ABC1}
	v_{i}=x_{i}+\varphi \times (x_{r1}-x_{r2}) ,
	\end{equation} 
	where $x_{r1}$ and $x_{r2}$ are two random candidate solutions and $\varphi$ is a random number from a uniform distribution in $[-1;1]$, and the better of $v_{i}$ and $x_{i}$ is kept. In the onlooker bee phase, a base candidate solution $x_{i}$ is selected based on the quality of each candidate solution. If the quality of an employed bee or onlooker does not improve over a number of trials, it converts into a scout and generates a random candidate solution.
\end{compactitem}
\subsubsection{Advanced Algorithms}

\begin{compactitem}
	\item \textbf{Levy-based Evolutionary Strategy (LevyES):} Levy flight is a specific kind of random walk that uses a Levy distribution to determine step size. The next position in a Markov chain that is called a "random walk" relies simply on the present position. A sequence generated by Levy flight involves a lot of little steps and occasionally big jumps. LevyES benefits from the random numbers generated by the Levy flight distribution rather than the uniform distribution, leading to more exploration and exploitation, simultaneously. 
	\item \textbf{Self-Adaptive DE (SADE)~\cite{SADE_DE}} is an improved variant of DE, based on the idea of employing two mutation operators, \textit{DE/rand/1} and \textit{DE /current-to-best/1}, simultaneously. \textit{DE /current-to-best/1} is defined as
	\begin{equation}
	\label{SADE}
	v_{i}=x_{i}+ F_{i} . (x_{best}-x_{i})+ F_{i} . (x_{r1}-x_{r2}),
	\end{equation}
	where $x_{best}$ is the best candidate solution from the current population, $x_{r1}$ and $x_{r2}$ are two randomly-selected candidate solutions, and $F_{i}$ is the scaling factor for $i$-th candidate solution. 
	
	\item 	\textbf{DE with Self-Adaptation Populations (SAP-DE)~\cite{SAP_DE}:} SAP-DE tries to present a self-adaptive population size in addition to self-adaptive crossover and mutation rates. To this end, SAP-DE proposes two variants, called SAP-DE-ABS and SAP-DE-REL to define a population size $\pi$. SAP-DE-ABS defines $\pi$ as
	\begin{equation}
	\label{Eq: SAP_DE}
	\pi = round (NP_{ini}+N(0,1))
	\end{equation}     
	while SAP-DE-REL initialises the population size parameter based on a uniform distribution between [-0.5,+0.5]. In each stage, the $\pi$ parameter should be updated. While SAP-DE-REL takes into account the current population size plus a percentage increase or decrease in accordance with the population growth rate, SAP-DE-ABS considers the population size of subsequent generations as the average of the population size attribute from all individuals in the current population. 
	\item \textbf{Adaptive DE with Optional External Archive (JADE)~\cite{JADE_01}} is a state-of-the-art variant of DE, defined based on three new modifications. First, JADE employs an \textit{archive}, including historical data, to select parents. Second, JADE introduces a novel mutation, \textit{DE /current-to- pbest}, as
	\begin{equation}
	\label{JADE}
	v_{i}=x_{i}+ F_{i} . (x_{best}^{p}-x_{i})+ F_{i} . (x_{r1}-x_{r2}),
	\end{equation} 
	where $x_{i}$ is the parent candidate solution, $x_{best}^{p}$ is a randomly-selected candidate solution from the best 100p\% candidate solutions in the current population, $x_{r1}$ and $x_{r2}$ are two candidate solutions randomly selected from the union of the current population and the archive. 
	
	The third modification is to select $F$ and $CR$, adaptively. JADE employs a normal distribution-based sampling for $CR$, while a Cauchy distribution is used to select $F$ values.   
	\item \textbf{Chaos PSO (CPSO)}~\cite{Chaos-PSO}: CPSO benefits from an adaptive inertia weight factor (AIWF) and a chaotic local search (CLS). AIWF leads to set $\omega$, in the original PSO, adaptively based on the objective function value as
	\begin{equation}
	\omega=\begin{cases}
	\omega_{min}+\frac{(\omega_{max}-\omega_{min})(f-f_{min})}{f_{avg}-f_{min}} & f \leq f_{avg} \\
	\omega_{max} & f \geq f_{avg}
	\end{cases} ,
	\end{equation}
	where $\omega_{max}$ and $\omega_{min}$ signify the minimum and maximum of $\omega$, respectively; $f$ is the current objective function value of a candidate solution, and $f_{avg}$ and $f_{min}$ are the average and minimum values of all candidate solutions, respectively. 
	
	To enhance the effectiveness, the CLS operator acts as a local search around the best position as
	\begin{equation}
	\label{CLS}
	cx_{i}^{k+1}=4cx_{i}^{k}(1-cx_{i})
	\end{equation}
	where $cx_{i}$ shows the $i$-th chaotic variable, and $k$ is the iteration number. $cx_{i}$ is distributed between 0 and 1, and the above equation shows a chaotic behavior when the initial  $cx_{0} \in (0,1)$ and $x_{0} \notin {0.25,0.5,0.75}$. 
	\item \textbf{Comprehensive Learning PSO (CLPSO)~\cite{CLPSO02}}: it suggests a comprehensive learning (CL) strategy for particle learning to avoid premature convergence. All particles's \textit{pbest} can be employed to adjust the velocity of each particle rather than just its own \textit{pbest}. The updating scheme in CLPSO is defined as
	\begin{equation}
	\label{Eq:vel2}
	v_{t+1}^{i}= \omega v_{t}^{i}+c_{1}  r (pbest_{fi(d)}^{f}-x_{t}^{i}),
	\end{equation}
	where $fi(d)$ defines which particles' \textit{pbest} particle $i$ should follow. The decision to learn from nearby particles is made using the comprehensive learning probability, $PC$. A random number with a uniform distribution is chosen for every dimension. The dependent dimension will learn from its own \textit{pbest} if the produced random number is greater than $P C(i)$. Otherwise, it updates based on the nearby particles.  
	
	\item \textbf{Self-organising Hierarchical PSO with Jumping Time-varying Acceleration Coefficients (HPSO)~\cite{HPSO-TVAC}}: the main characteristics of HPSO are as follows:
	\begin{enumerate}
		\item Mutation is defined for the PSO algorithm,
		\item A novel concept, called self-organizing hierarchical particle swarm optimiser with
		TVAC, is introduced which solely takes into account the "social" and "cognitive" components of the particle swarm strategy when estimating each particle's new velocity, and particles are re-initialised when they are stagnated in the search space. 
		\item A time-varying mutation step size is included in the PSO algorithm.
	\end{enumerate}
	\item \textbf{Phasor PSO(P-PSO)~\cite{Phasor-PSO}} proposes PSO control parameters based on a phase angle ($\theta$), inspired from phasor theory in mathematics. In each iteration, the velocity is updated as	
	\begin{equation}
	\label{Phasor-PSO}
	v_{i}^{iter}= |cos\theta_{i}^{iter}|^{2*sin\theta_{i}^{iter}} \times (Pbest_{i}^{iter}-x_{i}^{iter})
	+|sin\theta_{i}^{iter}|^{2*cos\theta_{i}^{iter}} \times (Gbest_{i}^{iter}-x_{i}^{iter})
	\end{equation}
	
\end{compactitem}

\subsubsection{Metaphor-based Algorithms}

\begin{compactitem}
	\item \textbf{Harmony Search (HS)~\cite{Harmony_original}}: it updates a new harmony (candidate solution) based on three rules, including, memory consideration, pitch adjustment and random selection. While random selection will explore the global search, and in consequence enhancing exploration, memory consideration and pitch adjustment ensure that the good local solutions are kept.
	\item\textbf{ Grey Wolf Optimiser (GWO)}~\cite{GWO_Main_Paper}: it is inspired by the social structure and hunting techniques of grey wolves. Based on the top three candidate solutions from the present population, each candidate solution is updated as
	\begin{equation}
	x_{i}(t+1)=(x_{1}+x_{2}+x_{3})/3 ,
	\end{equation}
	with
	\begin{equation}
	x_{1}=x_{\alpha}-r_{1} D_{\alpha} \mbox{  ,  }
	x_{2}=x_{\beta}-r_{2} D_{\beta} \mbox{  ,  }
	x_{3}=x_{\gamma}-r_{3} D_{\gamma} ,
	\end{equation}
	and
	\begin{equation}
	D_{\alpha}=|C_{1} x_{\alpha}-x_{i}(t)|  \mbox{  ,  }
	D_{\beta}=|C_{2} x_{\beta}-x_{i}(t)|  \mbox{  ,  }
	D_{\gamma}=|C_{3} x_{\gamma}-x_{i}(t)| ,
	\end{equation}
	where $x_{\alpha}$, $x_{\beta}$, and $x_{\gamma}$ are the best three candidate solutions, $r_{1}$, $r_{2}$, and $r_{3}$ are random numbers as are $D_{\alpha}$, $D_{\beta}$, and $D_{\gamma}$ and $C_{1}$, $C_{2}$, and $C_{3}$.
	
	\item \textbf{Ant Lion Optimiser (ALO)~\cite{ALO_Main_Paper}}: ALO is based on the hunting habits of ant lions. Six operators make up the basis of ALO, including the elitist approach, ant lion pit trapping, constructing traps, pushing ants toward the ant lion, and random walks of ants.
	
	\item \textbf{Dragonfly Algorithm (DA)~\cite{DA_Main_Paper}}: DA includes the following five factors: separation, alignment, cohesion, attraction, and distraction. While alignment is dependent on the mean of all neighbors' velocities, separation and cohesion depend on nearby individuals. The distance between the current candidate solution's location and the location of a food source determines attraction, whereas the distance between the current candidate solution's location and the location of an adversary determines distraction. Based on a combination of these five factors, each candidate solution is updated.
	
	\item \textbf{Whale Optimisation Algorithm (WOA)}~\cite{WOA_Main_Paper}: the social behavior of humpback whales is modeled by the WOA. It has three fundamental operators: encircling prey, a bubble-net attack, and spiral updating position. Encircling prey modifies the positions of each candidate solution dependent on the area around the best candidate solution. In the early rounds of the algorithm, the bubble-net attack involves a decreasing encircling motion that leads to exploration; in the later iterations, it leads to exploitation. It also involves updating depending on the distance between the current position and the best one.
	
	\item \textbf{Sine Cosine Algorithm (SCA)~\cite{SCA_Main_Paper}}: the behavior of the sine and cosine functions serves as the foundation for the SCA algorithm. Each candidate solution is updated as 
	\begin{equation}
	x_{i}(t+1)=\begin{cases}
	x_{i}(t)+r_{1} \sin(r_{2})+|r_{3} p_{i}(t)-x_{i}(t)| & \mbox{if \ } r_{4}<0.5 \\
	x_{i}(t)+r_{1} \cos(r_{2})+|r_{3} p_{i}(t)-x_{i}(t)| & \mbox{if \ } r_{4} \geq 0.5 
	\end{cases} ,
	\end{equation}
	where $p_{i}$ shows the destination solution, $r_{1}$ is a conversation parameter, $r_{2}$ a random number between 0 and $2\pi$, and $r3$ is a random number for weighing $p_{i}(t)$.
	
	\item \textbf{Gradient-based Optimiser (GBO)~\cite{GBO_Main_Paper}}: GBO as one of the most recent PBMHs is inspired by the gradient-based Newton’s method. The GBO algorithm tries to move based on a gradient-specified direction for each candidate solution in the current population.
	
	\item \textbf{Arithmetic Optimization Algorithm (AOA)~\cite{AOA_Main_Paper}:} AOA tries to find the optimal solution based on several arithmetic operators such as division and multiplication. The AOA algorithm benefits from two operators based on subtraction and addition for the exploitation phase, while the division search strategy and multiplication search strategy are responsible for the exploration phase.

\end{compactitem}

\section{Experimental Results}
\label{sec:Exp}
\begin{figure}[!htbp]
	\centering
	\begin{subfigure}[b]{0.25\linewidth}
		\includegraphics[width=\linewidth] {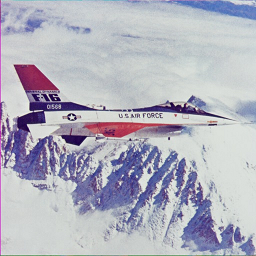}
		\caption{Airplane}
	\end{subfigure} 
	\begin{subfigure}[b]{0.25\linewidth}
		\includegraphics[width=\linewidth] {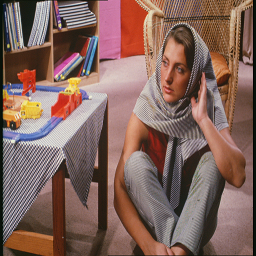}
		\caption{Barbara}
	\end{subfigure}
	\begin{subfigure}[b]{0.25\linewidth}
		\includegraphics[width=\linewidth] {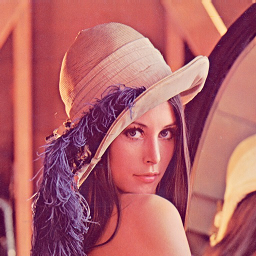}
		\caption{Barbara}
	\end{subfigure}
	\begin{subfigure}[b]{0.25\linewidth}
		\includegraphics[width=\linewidth] {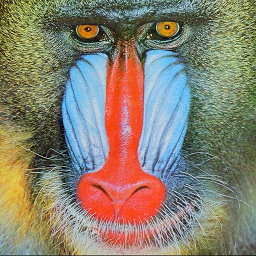}
		\caption{Mandrill}
	\end{subfigure} 
	\begin{subfigure}[b]{0.25\linewidth}
		\includegraphics[width=\linewidth] {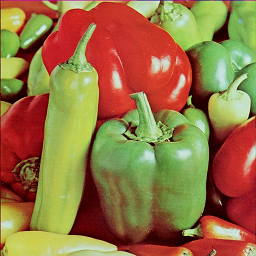}
		\caption{Peppers}
	\end{subfigure}
	\begin{subfigure}[b]{0.25\linewidth}
		\includegraphics[width=\linewidth] {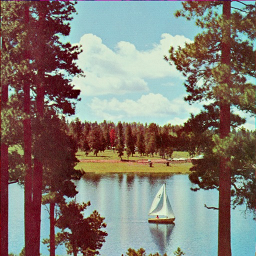}
		\caption{Sailboat}
	\end{subfigure} 
	\begin{subfigure}[b]{0.25\linewidth}
		\includegraphics[width=\linewidth] {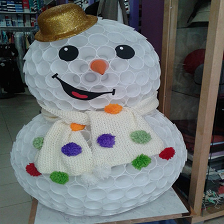}
		\caption{Snowman} 
	\end{subfigure} 	
	\begin{subfigure}[b]{0.25\linewidth}
		\includegraphics[width=\linewidth] {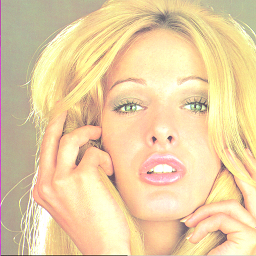}
		\caption{Tiffany}
	\end{subfigure} 
	\begin{subfigure}[b]{0.25\linewidth}
		\includegraphics[width=\linewidth] {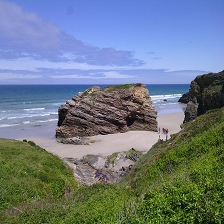}
		\caption{Beach}
	\end{subfigure}
	\begin{subfigure}[b]{0.25\linewidth}
		\includegraphics[width=\linewidth] {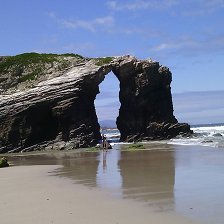}
		\caption{Cathedrals Beach}
	\end{subfigure} 
	\begin{subfigure}[b]{0.25\linewidth}
		\includegraphics[width=\linewidth] {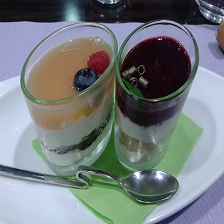}
		\caption{dessert}
	\end{subfigure}
	\begin{subfigure}[b]{0.25\linewidth}
		\includegraphics[width=\linewidth] {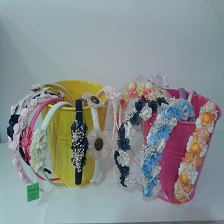}
		\caption{headbands}
	\end{subfigure}
	\begin{subfigure}[b]{0.25\linewidth}
		\includegraphics[width=\linewidth] {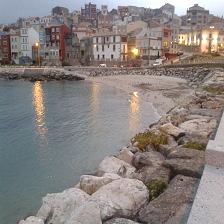}
		\caption{landscape}
	\end{subfigure} 
	
	\caption{Benchmark images.}
	\label{fig:bench}
\end{figure}

In this section, an extensive set of experiments is offered to show the effectiveness of our proposed strategy. To achieve this, we used 6 of the images recommended in \cite{color_quantisation_SFLA} for benchmarking image quantisation, including Snowman, Beach, Cathedrals Beach, Dessert, Headbands, and Landscape, in addition to 7 widely used benchmark images for image compression, including Airplane, Barbara, Lena, Mandrill, Peppers, Tiffany, and Sailboat. The benchmark images are displayed in Figure~\ref{fig:bench}.

Our proposed strategy is embedded in 22 PBMH-based search strategies. We categorised them into three main groups: base, advanced, and metaphor-based algorithms. Base algorithms are the most famous algorithms, while advanced algorithms are the state-of-the-art variants of the base algorithms. Also, we selected some metaphor-based algorithms according to newness and the number of citations. Some metaphor-based algorithms such as AOA have been presented in recent years, while others such as GWO have attracted a significant number of citations in recent years.  

To provide for a fair comparison, each algorithm is executed 30 times, independently. We provide the results for two user-specified file sizes, 10000 and 50000 bytes. For each algorithm, the population size and the number of function evaluations are set to 20 and 1000, respectively. Other parameters are set to their default values (Table~\ref{tab:parameter}) in the paper's appendix. All algorithms are implemented in Python and with the Mealpy framework~\cite{Mealpy}, one of the largest Python modules for the most cutting-edge metaheuristic algorithms.

\subsection{Evaluation Criteria}
We used the mean objective function, closeness, and confidence factor (CF) as the main criteria for evaluation.  Closeness refers to the distance between the size of the output image and the user-specified file size, defined as
\begin{equation}
C=|FS_{output} - FS_{US}|
\end{equation}
where $FS_{output}$ is the output of file size and $FS_{US}$ means user-specified file size. Lower closeness points out a higher ability of an algorithm to find an output image with a file size similar to the user-specified file size.

Also, we defined the CF measure as
\begin{equation}
CF =\frac{\sum_{ind=0}^{Nr} \xi }{Nr} 
\end{equation}

where $Nr$ is the number of independent runs for a specific algorithm (here is 30), CF is the confidence factor, and $\xi$ is defined as 
\begin{equation}
\xi=\begin{cases}
1 & \text{if } C<CS \\
0 & \text{otherwise}
\end{cases} ,
\end{equation}

where $C$ is the closeness measure, and $CS$ is called confidence coefficient, which shows an acceptable file size for output image (here CS is 10000). So a good algorithm will have a large value for CF, ideally it would be 1. 

In addition, for further clarification of the behavior of the search strategies for proposed approach, we provide three criteria, including diversity, exploitation, and exploration. 


\subsection{Results for the Base Algorithms}
The results for the base algorithms can be seen in Tables~\ref{tab:obj_base} to \ref{tab:acc_base} in the paper's appendix. In each table, we also provide the rank of each algorithm per image. From Table~\ref{tab:obj_base} in the paper's appendix, GA obtains the lowest rank in 24 out of 26 cases, and the second rank in only 2 cases, leading to the lowest average rank, and subsequently the first overall rank. The second average rank goes to ABC, while the MA algorithm ranks third. On the other hand, DE and PSO give the worst results. 

From Table~\ref{tab:diff_base} in the paper's appendix, we can observe that the results are promising. Some algorithms such as GA and ES are able to accurately find the image with the user-specified file size; for instance, the difference between the output file size for Barbara image is 23.43 and 40.03 bytes for the $FS_{US}=10000$ and $50000$, respectively. Based on the closeness measure, ES is the best-performing algorithm, followed by GA and ABC.

The results of the CF measure in Table~\ref{tab:acc_base} in the paper's appendix are consistent with the earlier tables, so that GA, ES, and ABC are the best algorithms. In particular, GA obtains the CF measure equal to 1 for all but four cases, indicating that in the majority of cases, it can find the desired image size. Some algorithms such as PSO, MA, and DE can not provide satisfactory results, for instance, the CF measure of the PSO algorithm is between 0.13 and 0.67 in all cases.

\subsection{Results for the Advanced Algorithms}

Advanced algorithms include the state-of-the-art variant of the base algorithms. To this end, we selected 8 algorithms as the search strategies. The mean objective function value and its rank for all algorithms and images are provided in Table~\ref{tab:obj_adv} in the paper's appendix. From the table, we can observe that HPSO performs best, followed by CLPSO (at a narrow margin) and JADE. The worst algorithms are SAP-DE, SADE, and LevyES. 

HPSO again obtained the lowest closeness measure (based on Table~\ref{tab:distance_adv} in the paper's appendix, while the second and third ranks belong to LevyES and CLPSO. It is worthwhile to mention that LevyES can not perform well in terms of objective function, but it is able to provide better results in terms of closeness measure. Again, SADE and SAP-DE yield the worst overall rank.  

Despite the efficiency of the HPSO algorithm in terms of objective function and closeness measures, HPSO has not been able to maintain its efficiency according to CF measure (Table~\ref{tab:acc_advanced} in the paper's appendix) and it is located in the third place, while the first and second ranks go to CPSO and CLPSO, respectively. Again, SADE and SAP-DE have the lowest rankings.

All in all, we can say that HPSO and CLPSO perform best among other advanced algorithms since HPSO has two first-place and one third-place overall ranks, while two second-place ranks and one third-overall-rank are obtained by CLPSO. 

\subsection{Results for the Metaphor-based Algorithms}
The experimental results for the metaphor-based algorithms are presented in Tables~\ref{tab:obj_metaphore} to \ref{tab:acc_metaphor} in the paper's appendix. From Table~\ref{tab:obj_metaphore} in the paper's appendix, it is possible to see that the WOA algorithm achieves the first overall rank, followed by GWO and HS. The interesting point is that while the AOA algorithm is one of the most recently introduced PBMH algorithms, it performs the worst. 

The objective function results are almost consistent with the closeness measure. From Table~\ref{tab:distance_metaphor} in the paper's appendix, The HS algorithm provides the lowest rank, while the second rank goes to the WOA algorithm. Also, SCA, DA and AOA offer the worst results. In some cases, none of metaphor-based algorithms can provide satisfactory results; for instance, for the Airplane image and $FS_{US}=10000$, the  closeness is between 184.33 and 1474.53. In some other cases, the effectiveness of the search strategy is more tangible; for instance for the Headbands image and with $FS_{US}=10000$, the HS algorithm achieves closeness equal to 5.83 (an impressive result), while SCA obtains 2898.27, demonstrating the effect of the search strategy.

The CF results for metaphor-based algorithms can be seen in Table~\ref{tab:acc_metaphor} in the paper's appendix. It is clear that WOA achieves the first average rank, followed by HS and GWO. Again, the worst ranks belong to SCA, AOA, and DA. By taking look at the table, some algorithms such as WOA yield satisfactory results by a CF of more than 0.97 in all cases, while some others such as SCA, can not achieve an accuracy higher than 0.5, indicating the high impact of the search strategy in the effectiveness of the proposed approach.

Finally, we can say that among metaphor-based algorithms, WOA, GA, and HS (as one of the oldest algorithms) presented the best performance, while DA, SCA, and AOA (as one of the recent algorithms) can be categorised as the worst algorithms.

\subsection{Overall Evaluation}
This section aims to provide an overall comparison among all 22 search strategies. To this end, we provide an overall ranking in Table~\ref{tab:overall} based on three criteria, objective function, closeness, and confidence factor. 

Based on the objective function, HPSO achieved the first rank, followed by CLPSO as second, and WOA as third. Only GA achieved a satisfactory rank among the base algorithms, while SAP-DE failed to achieve an acceptable rank among the advanced algorithms. The worst algorithms are DA, SAP-DE, and SCA algorithms.

HPSO obtained the second rank, based on the closeness measure, while the first rank belongs to the HS algorithm. It is worthwhile to mention that HS achieves the sixth rank based on the objective function. WOA again attains the third rank in terms of closeness measure.

CPSO is ranked first in terms of CF measure, while it can not achieve a low rank based on other criteria. The second and third ranks go to GA and CLPSO.

By moving from ES to LevyES, we can observe that LevyES can achieve a better performance in terms of objective function (14 to 11), and similar ranks based on the other two measures. A comparison between DE and their variants (SADE, SAP-DE, and JADE) reveals that SADE and JADE can improve the results of the DE algorithm, while  while SAP-DE fails to do so. In particular, the JADE algorithm performs best among other DE variants. In addition, a comparison between PSO and its variants, including CPSO, CLPSO, HPSO, and PPSO shows that all variants outperform the standard PSO algorithm, and in particular, HPSO obtained excellent results.

Another point that can be found using the table is that some algorithms have shown great diversity in different criteria. For instance, closeness for CPSO shows a rank of 11, while its CF is 1. It shows that the mean file size obtained by CPSO is higher than some algorithms, but in most cases, CF is better or the file size is lower than the threshold.  From the table, GA, CLPSO, HPSO, HS, and WOA can overcome others, and the results benefit from a lower diversity.
\begin{table}[tb]
	\centering
	\caption{Overall ranking of the algorithms based on objective function, closeness, and accuracy.}
	\label{tab:overall}
	\begin{tabular}{l|l|ccc}
		\hline
		Category                                   & Algorithms & Objective function & Closeness & CF \\ \hline
		\multirow{6}{*}{Base Algorithms}           & GA         & 5                  & 6    & 2        \\
		& DE         & 17                 & 20   & 21       \\
		& MA         & 16                 & 16   & 16       \\
		& PSO        & 19                 & 22   & 22       \\
		& ES         & 15                 & 4    & 9.5      \\
		& ABC        & 14                 & 13   & 11       \\ \hline
		\multirow{8}{*}{Advanced Algorithms}       & LevyES     & 11                 & 5    & 9.5      \\
		& SADE       & 12                 & 12   & 13       \\
		& SAP-DE    & 21                 & 17   & 17       \\
		& JADE       & 6                  & 10   & 5        \\
		& CPSO       & 8                  & 11   & 1        \\
		& CLPSO      & 2                  & 7    & 3        \\
		& HPSO       & 1                  & 2    & 4        \\
		& PPSO       & 10                 & 9    & 12       \\ \hline
		\multirow{8}{*}{Metaphor-based algorithms} & HS         & 6                  & 1    & 6        \\
		& GWO        & 4                  & 8    & 8        \\
		& ALO        & 9                  & 15   & 15      \\
		& DA         & 22                 & 19   & 18      \\
		& WOA        & 3                  & 3    & 7        \\
		& SCA        & 20                 & 21   & 20       \\
		& GBO        & 13                 & 14   & 14       \\
		& AOA        & 18                 & 18   & 19        \\ \hline
	\end{tabular}
\end{table}  

For further explanation, we also provide the results of Wilcoxon signed rank test at 5\% significance level, as a pairwise statistical test, between all combinations of the algorithms to show whether the algorithms are significantly different or not. 

The test shows whether there is a statistically significant difference between the performance of any two algorithms. While the alternative hypothesis, known as $H_{1}$, suggests that there is a noticeable difference, the $H_{0}$ hypothesis exposes the same behavior of the two algorithms under evaluation. The significance level shows the rejection of probability of $H_{0}$. If the computed $p$-value is less than the significance level, $H_{0}$ is rejected. The results of the statistical tests are shown in Table~\ref{tab:wilc}.

From the table, we can see that HPSO performs significantly better than the others; in 19 cases, HPSO outperforms others significantly, while the results of HPSO are equivalent to PSO and ES. The overall next best performing algorithms are CLPSO and WOA (18 wins, 2 ties, and 1 fail), and GWO (18 wins and 3 fails). On the other hand, several algorithms such as DA (21 fails), SAP-DE (1 win, 2 ties, and 18 fails), and SCA (2 wins, 1 tie, and 18 fails) can not perform well enough in comparison to others.

\begin{sidewaystable}[!htbp]
	\setlength{\tabcolsep}{2pt}
	\centering
	\caption{ Results of Wilcoxon signed rank test based on mean objective function value. $+$, $-$, and $=$ denote that the algorithm in the corresponding row is statistically superior than, inferior to, or equivalent to the algorithm in the corresponding column. The last column includes a summary of the number of wins ($w$), ties ($t$) and losses ($l$) for the algorithms.}
	\label{tab:wilc}
	\small
	\begin{tabular}{l|cccccccccccccccccccccc|c}
		\hline
		\hline
		& GA & DE & MA & PSO & ES & ABC & LevyES & SADE & SAP-DE & JADE & CPSO & CLPSO & HPSO & PPSO & HS & GWO & ALO & DA & WOA & SCA & GBO & AOA & \textit{w/t/l}   \\
		\hline
		\hline
		GA      & \cellcolor{gray!10}  & +  & +  & +   & +  & +   & +      & +    & +       & +    & +    & -     & -    & +    & +  & -   & +   & +  & -   & +   & +   & +   & 17/0/4  \\
		DE      & -  & \cellcolor{gray!10}  & -  & +   & -  & -   & -      & -    & =       & =    & -    & -     & -    & -    & -  & -   & -   & +  & -   & +   & -   & +   & 4/2/15  \\
		MA      & -  & +  &  \cellcolor{gray!10}  & +   & -  & -   & -      & -    & +       & -    & -    & =     & -    & =    & -  & -   & -   & +  & -   & +   & -   & +   & 6/2/13  \\
		PSO     & -  & -  & -  &  \cellcolor{gray!10}  & =  & -   & -      & -    & +       & -    & -    & -     & =    & -    & -  & -   & -   & +  & -   & +   & -   & -   & 3/2/16  \\
		ES      & -  & +  & +  & =   &  \cellcolor{gray!10}  & -   & -      & -    & +       & -    & -    & -     & =    & -    & -  & -   & -   & +  & -   & +   & -   & +   & 6/2/13  \\
		ABC     & -  & +  & +  & +   & +  & \cellcolor{gray!10}   & -      & -    & +       & -    & -    & -     & -    & -    & -  & -   & -   & +  & -   & +   & -   & +   & 8/0/13  \\
		LevyES  & -  & +  & +  & +   & +  & +   &  \cellcolor{gray!10}   & +    & +       & -    & -    & -     & -    & -    & -  & -   & -   & +  & -   & +   & +   & +   & 11/0/10 \\
		SADE    & -  & +  & +  & +   & +  & +   & -      &  \cellcolor{gray!10}   & +       & -    & -    & -     & -    & =    & -  & -   & -   & +  & -   & +   & +   & +   & 10/1/10   \\
		SAP-DE & -  & =  & -  & -   & -  & -   & -      & -    &  \cellcolor{gray!10}      & =    & -    & -     & -    & -    & -  & -   & -   & +  & -   & -   & -   & -   & 1/2/18  \\
		JADE    & -  & =  & +  & +   & +  & +   & +      & +    & =       &  \cellcolor{gray!10}     & +    & -     & -    & +    & +  & -   & +   & +  & -   & +   & +   & +   & 14/2/5  \\
		CPSO    & -  & +  & +  & +   & +  & +   & +      & +    & +       & -    &  \cellcolor{gray!10}    & -     & -    & +    & -  & -   & +   & +  & -   & +   & =   & +   & 13/1/7  \\
		CLPSO   & +  & +  & =  & +   & +  & +   & +      & +    & +       & +    & +    &  \cellcolor{gray!10}    & -    & =    & +  & +   & +   & +  & +   & +   & +   & +   & 18/2/1  \\
		HPSO    & +  & +  & +  & =   & =  & +   & +      & +    & +       & +    & +    & +     &   \cellcolor{gray!10}   & +    & +  & +   & +   & +  & +   & +   & +   & +   & 19/2/0  \\
		PPSO    & -  & +  & =  & +   & +  & +   & +      & =    & +       & -    & -    & =     & -    & \cellcolor{gray!10}  & -  & -   & -   & +  & -   & +   & +   & +   & 10/3/8  \\
		HS      & -  & +  & +  & +   & +  & +   & +      & +    & +       & -    & +    & -     & -    & +    & \cellcolor{gray!10}   & -   & +   & +  & -   & +   & +   & +   & 15/0/6  \\
		GWO     & +  & +  & +  & +   & +  & +   & +      & +    & +       & +    & +    & -     & -    & +    & +  &  \cellcolor{gray!10}   & +   & +  & -   & +   & +   & +   & 18/0/3  \\
		ALO     & -  & +  & +  & +   & +  & +   & +      & +    & +       & -    & -    & -     & -    & +    & -  & -   &  \cellcolor{gray!10}  & +  & -   & +   & +   & +   & 13/0/8  \\
		DA      & -  & -  & -  & -   & -  & -   & -      & -    & -       & -    & -    & -     & -    & -    & -  & -   & -   &  \cellcolor{gray!10}  & -   & -   & -   & -   & 0/0/21    \\
		WOA     & +  & +  & +  & +   & +  & +   & +      & +    & +       & +    & +    & -     & -    & +    & +  & +   & +   & +  &  \cellcolor{gray!10}  & =   & +   & +   & 18/1/2  \\
		SCA     & -  & -  & -  & -   & -  & -   & -      & -    & +       & -    & -    & -     & -    & -    & -  & -   & -   & +  & =   & \cellcolor{gray!10}  & -   & -   & 2/1/18  \\
		GBO     & -  & +  & +  & +   & +  & +   & -      & -    & +       & -    & =    & -     & -    & -    & -  & -   & -   & +  & -   & +   & \cellcolor{gray!10}  & +   & 9/1/11  \\
		AOA     & -  & -  & -  & +   & -  & -   & -      & -    & +       & -    & -    & -     & -    & -    & -  & -   & -   & +  & -   & +   & -   &  \cellcolor{gray!10}  & 4/0/17 \\
		\hline
		\hline
	\end{tabular}
\end{sidewaystable}

\subsection{Further Discussion}
\begin{figure}[tb]
	\centering
	\includegraphics[width=1\columnwidth]{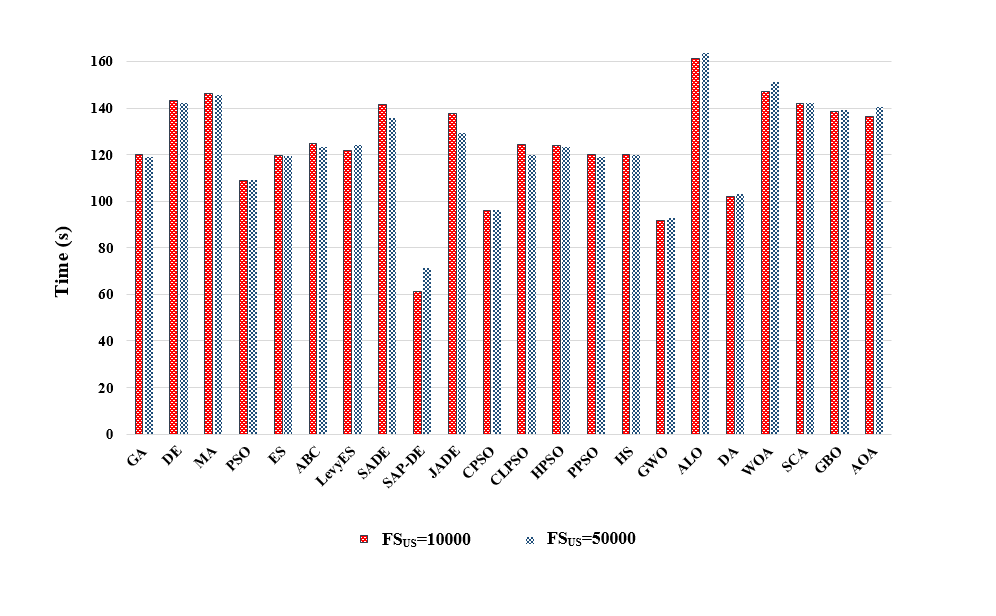}
	\caption{Computation time for different algorithms and the Airplane image.} 
	\label{fig:Time}
\end{figure}

This section provides a further discussion on the behaviour analysis of the search strategies, in particular, based on computation time, population diversity, exploration, and exploitation. In the first experiment, we evaluate the algorithms regarding computation time. It is worth mentioning that the representation and the objective function are the same for all algorithms. Therefore, the amount of computation time differences are related to the search strategy. The experiments were run using a Desktop PC with Linux Version 42.2, a i7-7700k CPU at 4.20GHz, 64GB RAM, and a 1 TB SSD hard. Figure~\ref{fig:Time} shows the average computation time for all algorithms and a representative image, \textit{Airplane}. At first sight, we can observe that the computation time for most algorithms is between a little less than 100 seconds and a little more than 140 seconds. The exceptions are SAP-DE and ALO. The SAP-DE algorithm took the least time to run, but based on the earlier results, it has yet to be able to perform well enough based on other criteria. Another finding is that for different file size and a same algorithm, the computation times are almost the same, meaning that different values for the user-specified file size have a small impact on the computation time.

In the next experiment, we have evaluated the population diversity in the population. Population diversity serves as a gauge for solution distribution in the population. In this study, we define population diversity using the Euclidean distance metric, as 

\begin{equation}
D=\frac{1}{NP} \sum_{i=1}^{NP}\sqrt{\sum_{k=1}^{D}(x_{ik}-\bar{x_{k}})}
\end{equation}

where $NP$ is the population size and $D$ is the problem dimension, $x_{ik}$ shows the $k$-th dimension of $i$-th individual, and $\bar{x_{k}}$ is the population mean, defined as

\begin{equation}
\bar{x_{k}} = \frac{1}{NP} \sum_{i=1}^{NP} x_{ik}
\end{equation}

given $\bar{x}=[x_{1},x_{2},...,x_{k},...,x_{D}]$.

\begin{figure}[tb!]
	\centering
	\begin{subfigure}[b]{0.30\linewidth}
		\includegraphics[width=\linewidth] {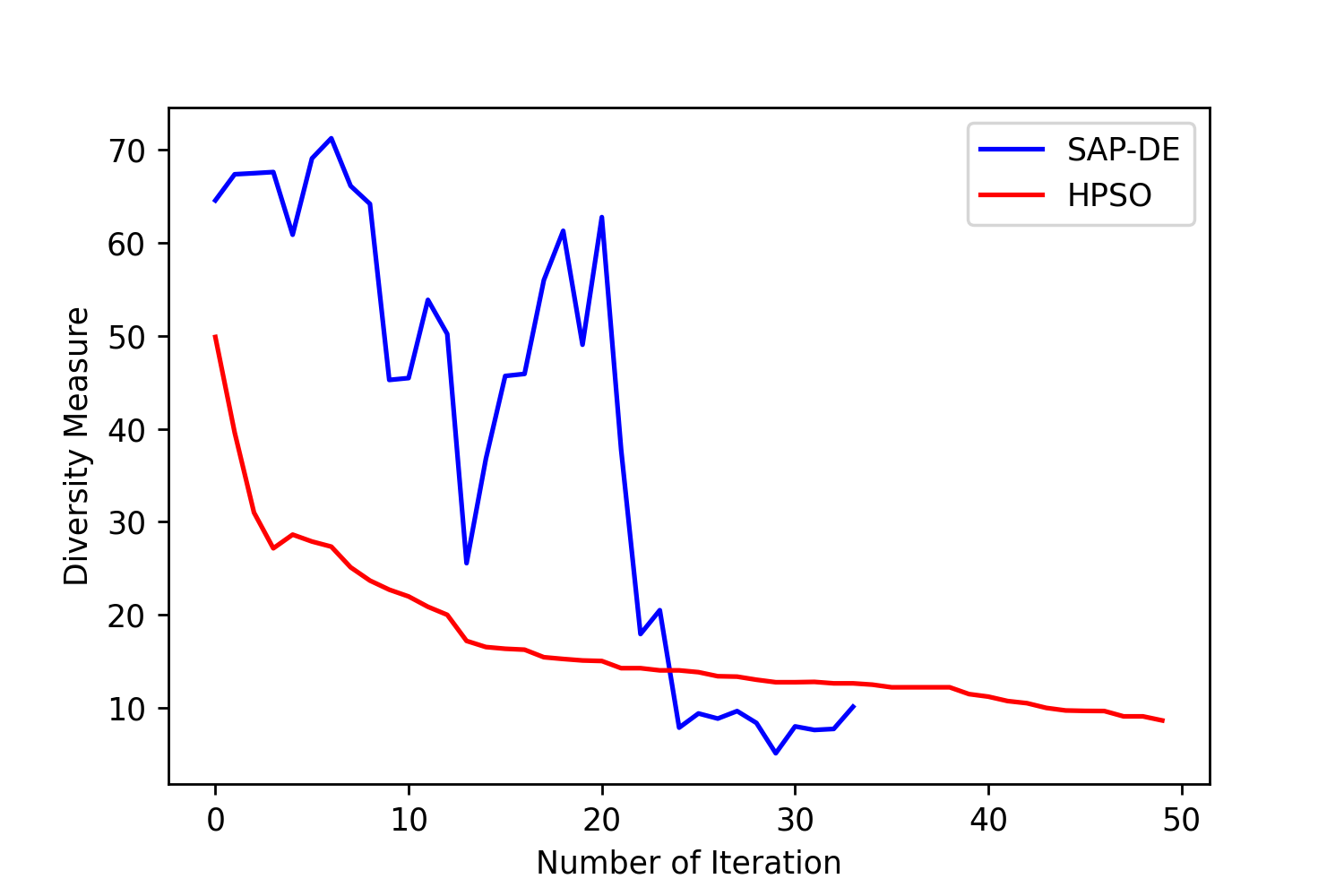}
		\caption{Airplane}
	\end{subfigure} 
	\begin{subfigure}[b]{0.30\linewidth}
		\includegraphics[width=\linewidth] {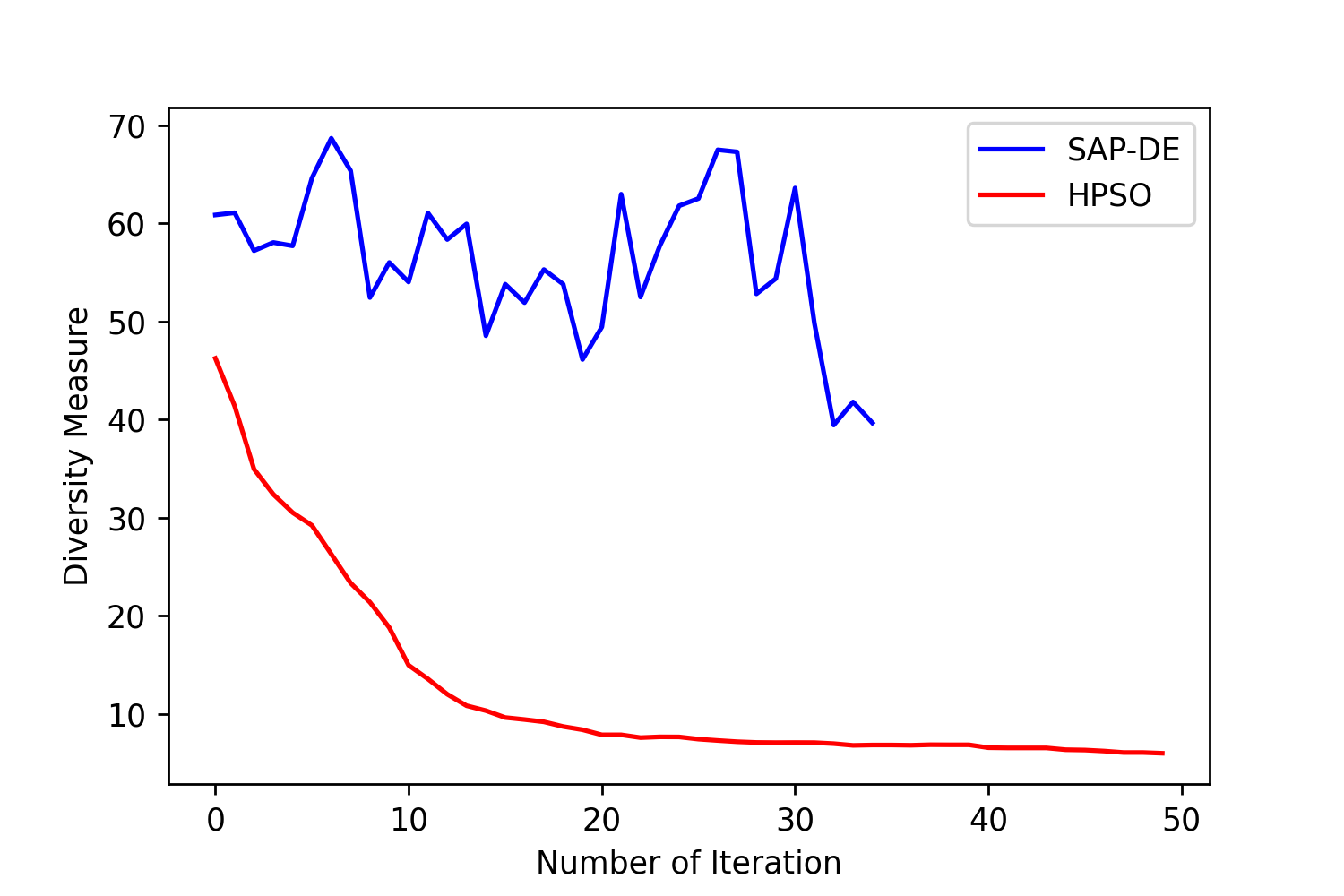}
		\caption{Barbara}
	\end{subfigure}
	\begin{subfigure}[b]{0.30\linewidth}
		\includegraphics[width=\linewidth] {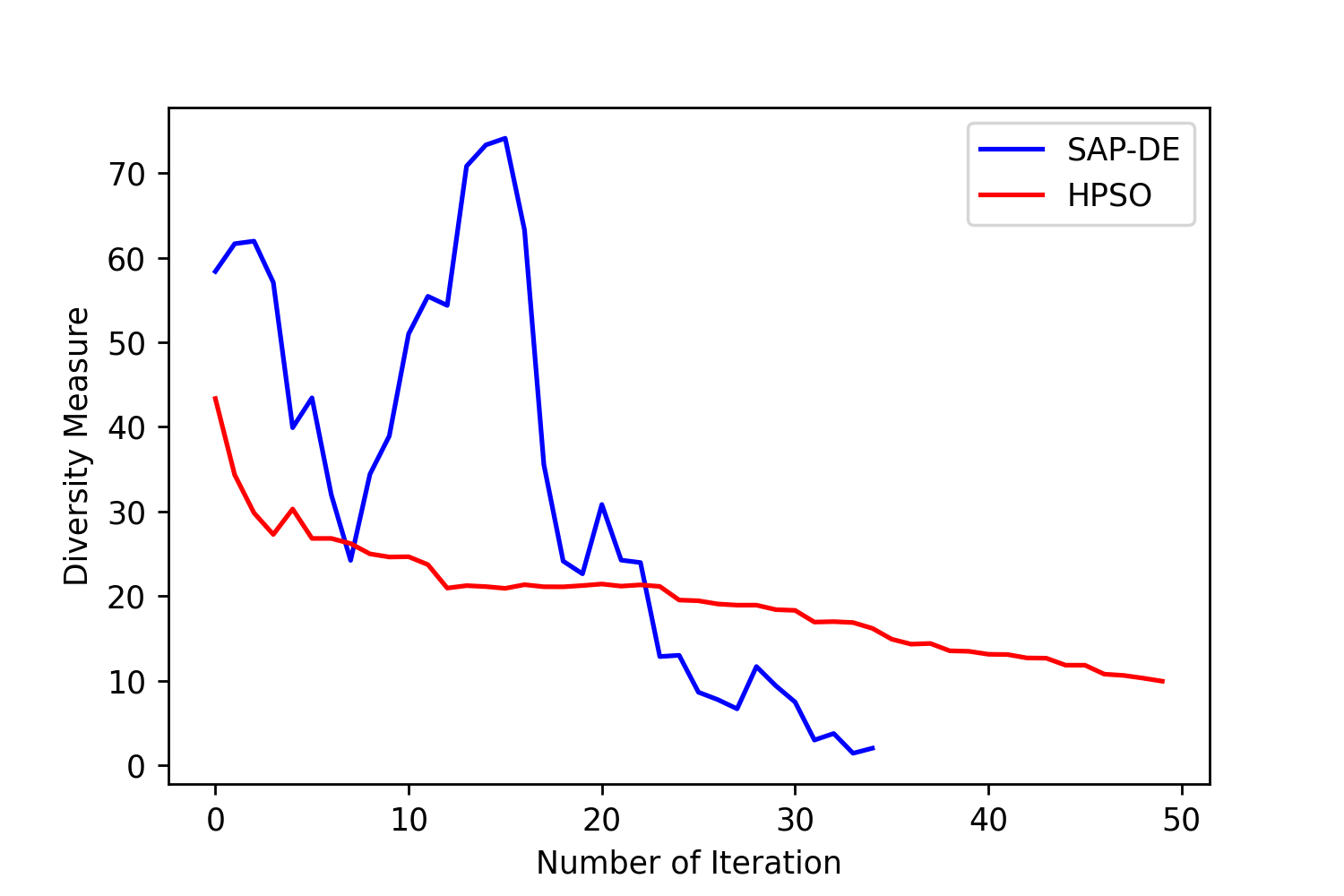}
		\caption{Barbara}
	\end{subfigure}
	\begin{subfigure}[b]{0.30\linewidth}
		\includegraphics[width=\linewidth] {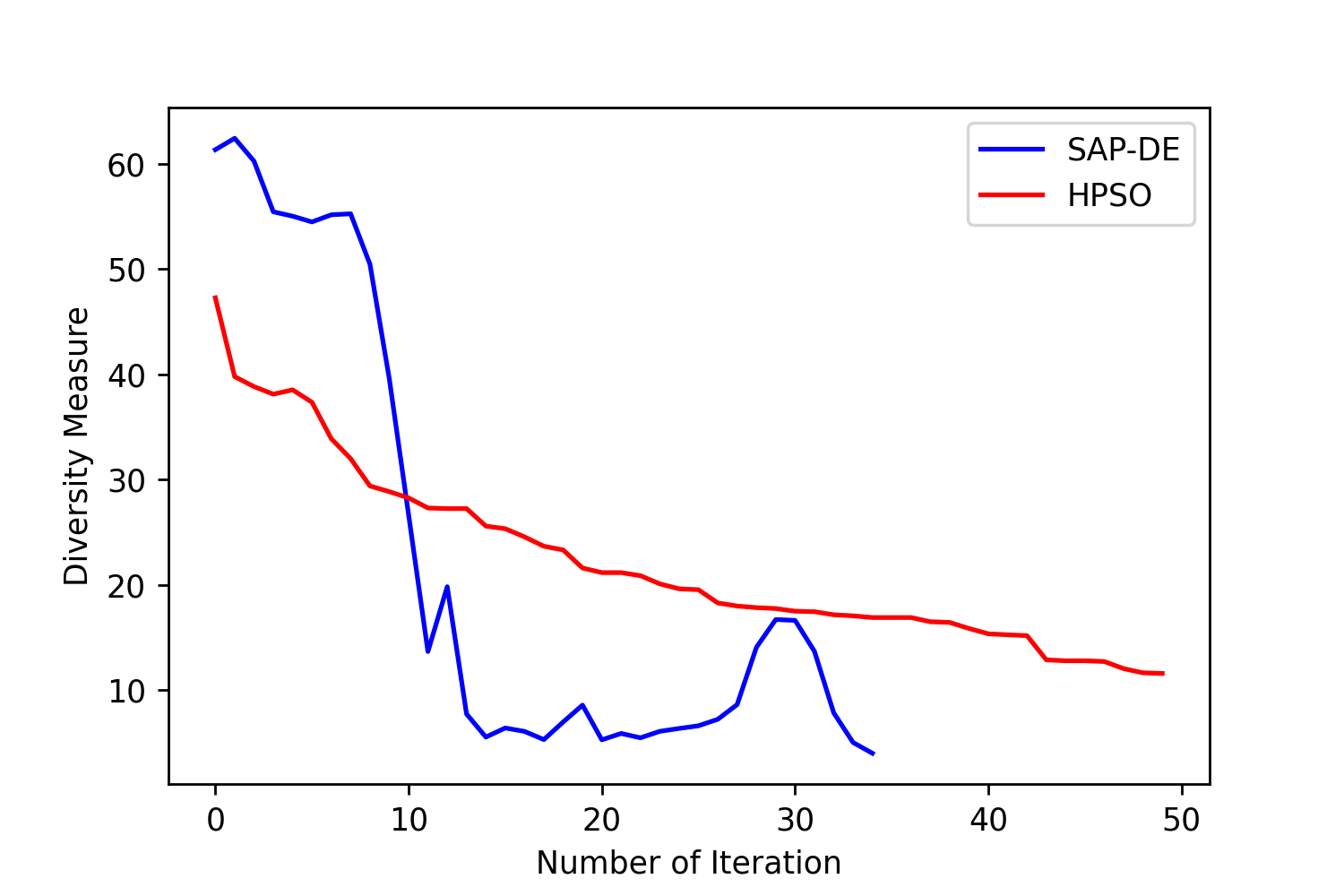}
		\caption{Mandrill}
	\end{subfigure} 
	\begin{subfigure}[b]{0.3\linewidth}
		\includegraphics[width=\linewidth] {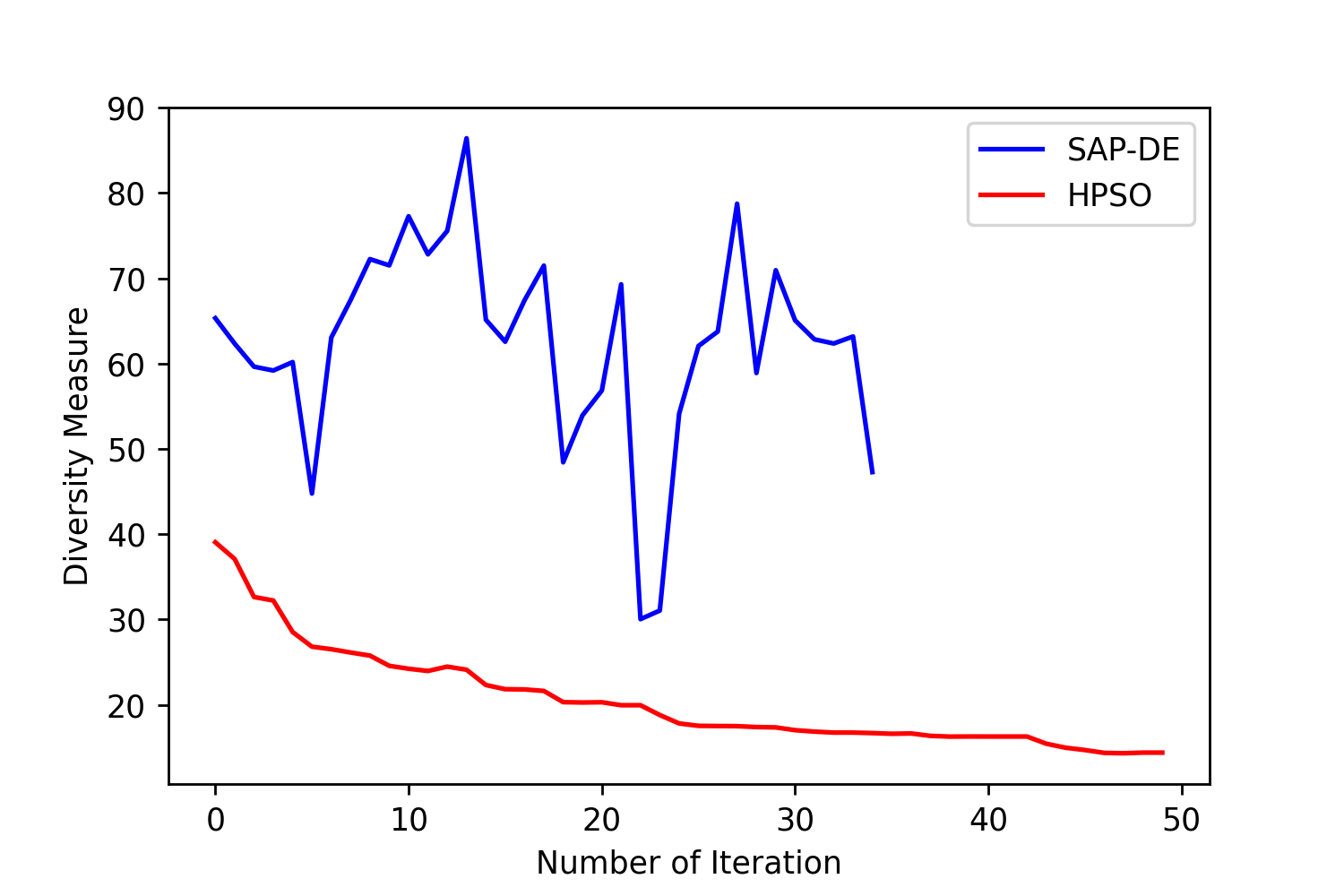}
		\caption{Peppers}
	\end{subfigure}
	\begin{subfigure}[b]{0.3\linewidth}
		\includegraphics[width=\linewidth] {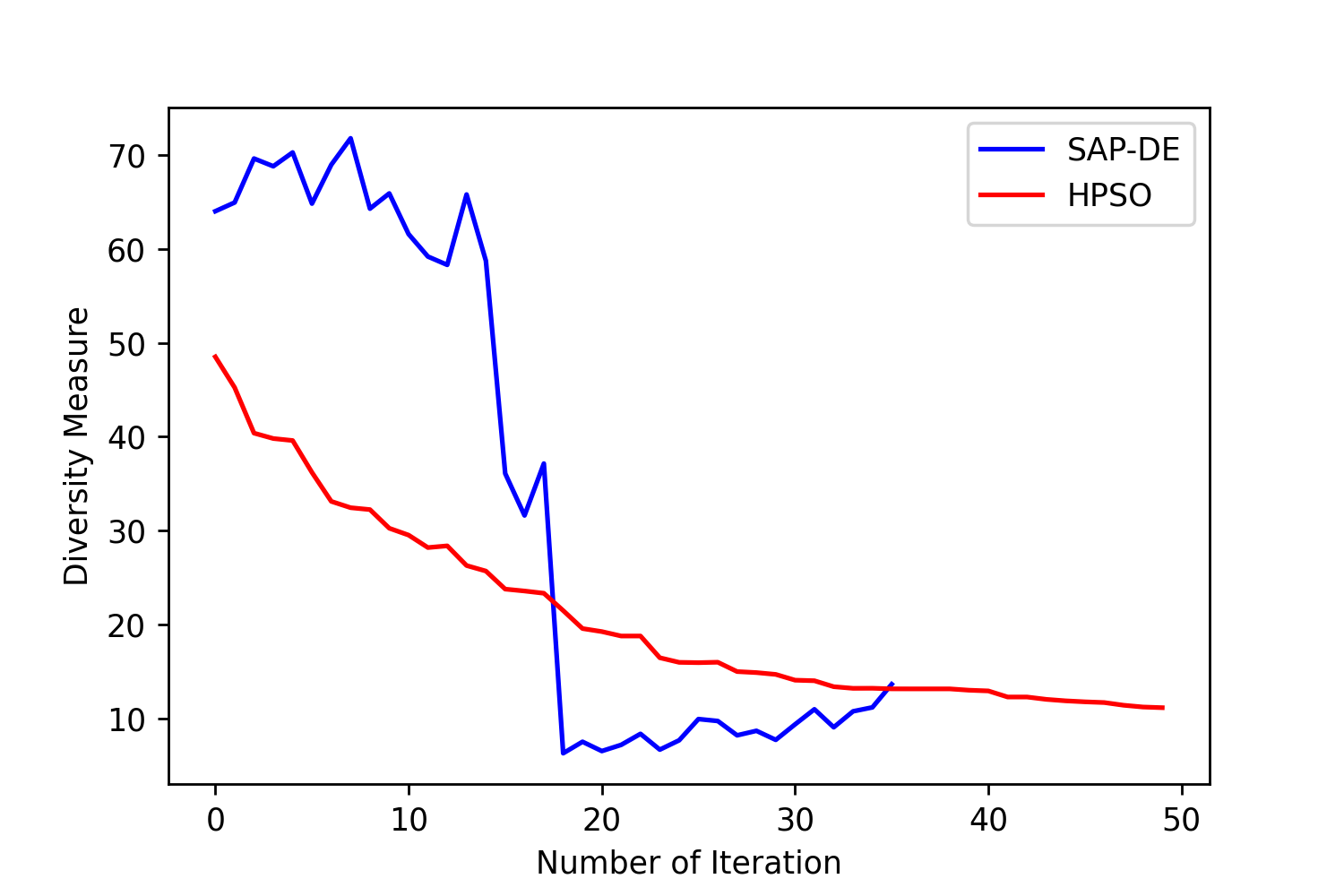}
		\caption{Sailboat}
	\end{subfigure} 
	\begin{subfigure}[b]{0.3\linewidth}
		\includegraphics[width=\linewidth] {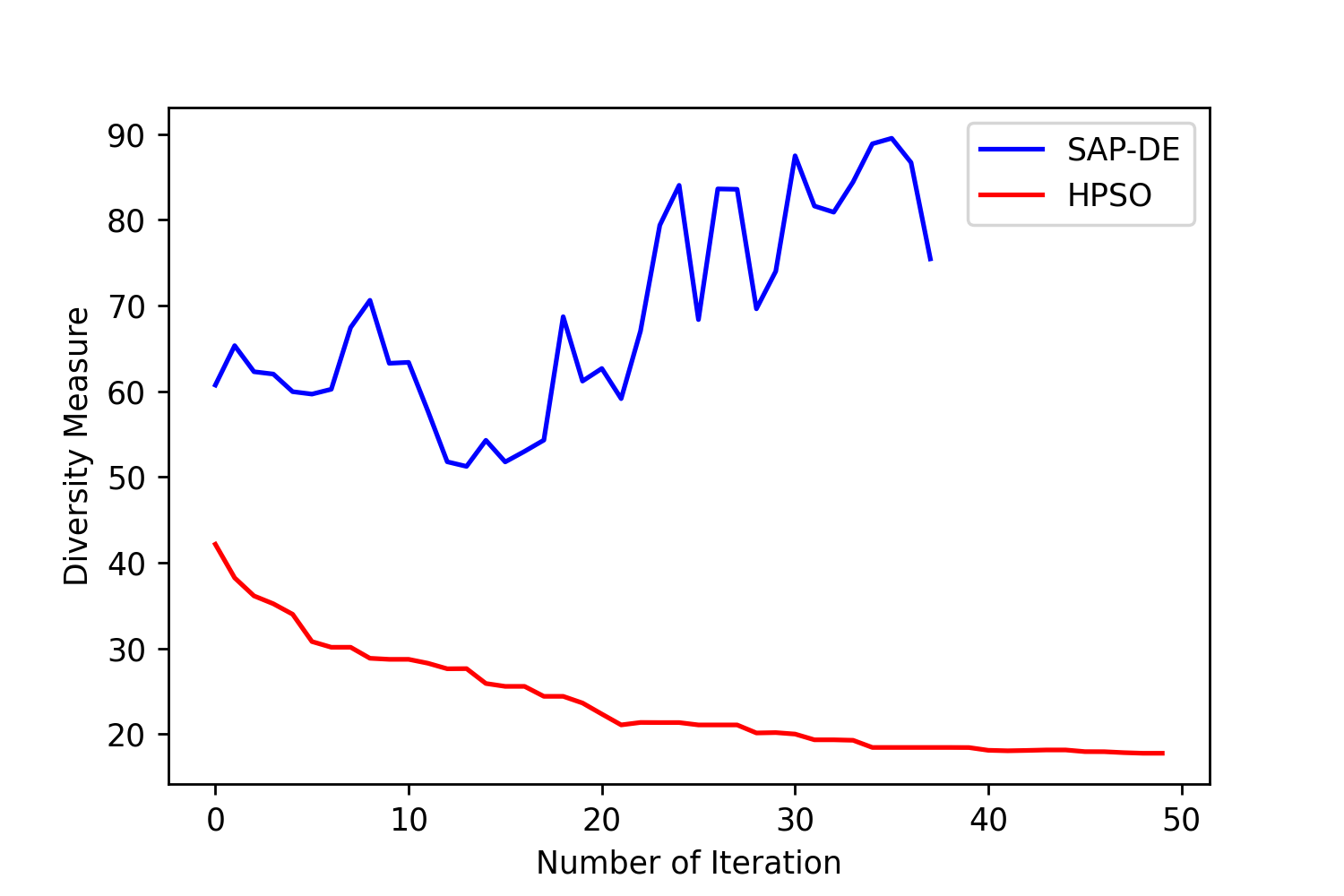}
		\caption{Snowman} 
	\end{subfigure} 	
	\begin{subfigure}[b]{0.3\linewidth}
		\includegraphics[width=\linewidth] {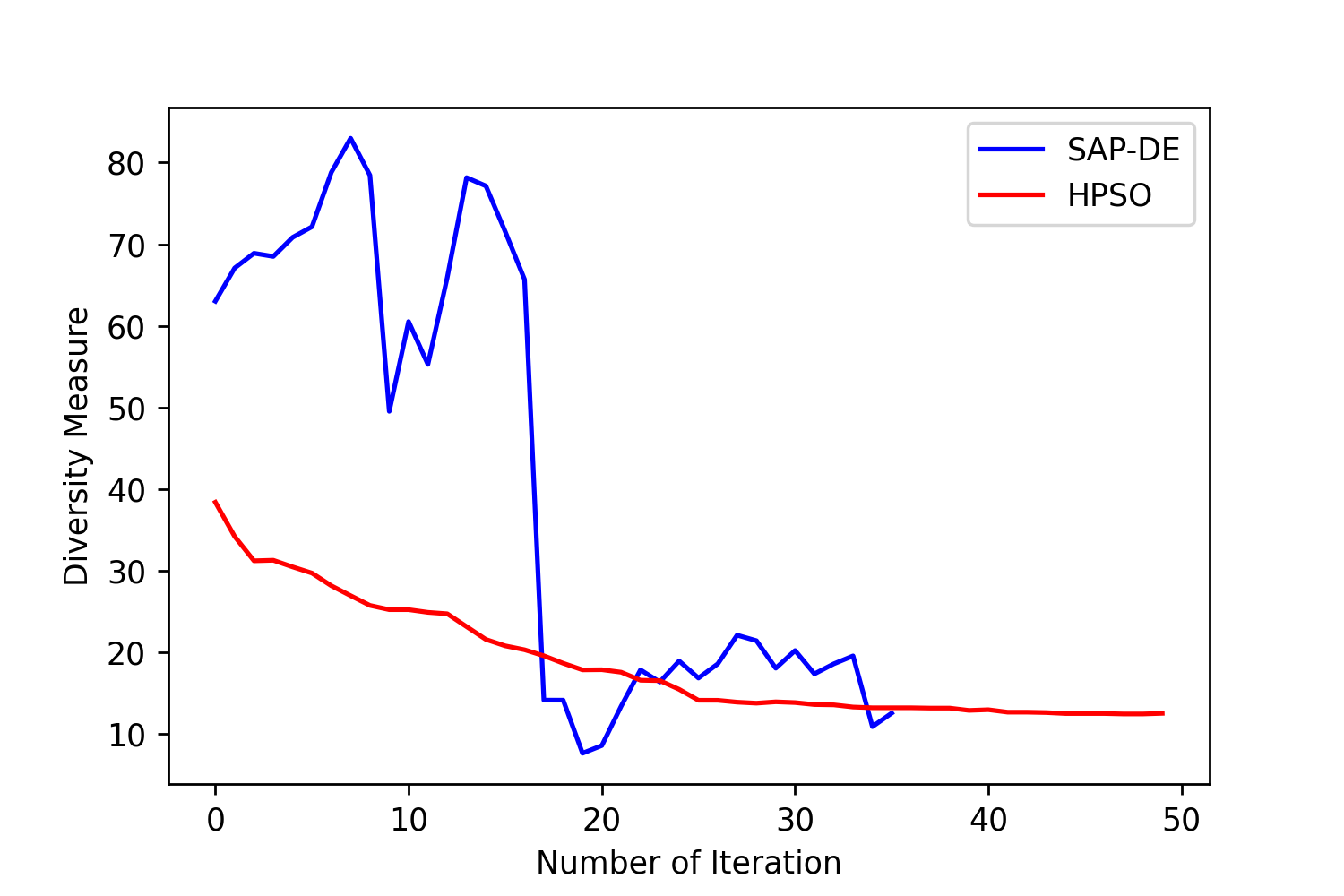}
		\caption{Tiffany}
	\end{subfigure} 
	\begin{subfigure}[b]{0.3\linewidth}
		\includegraphics[width=\linewidth] {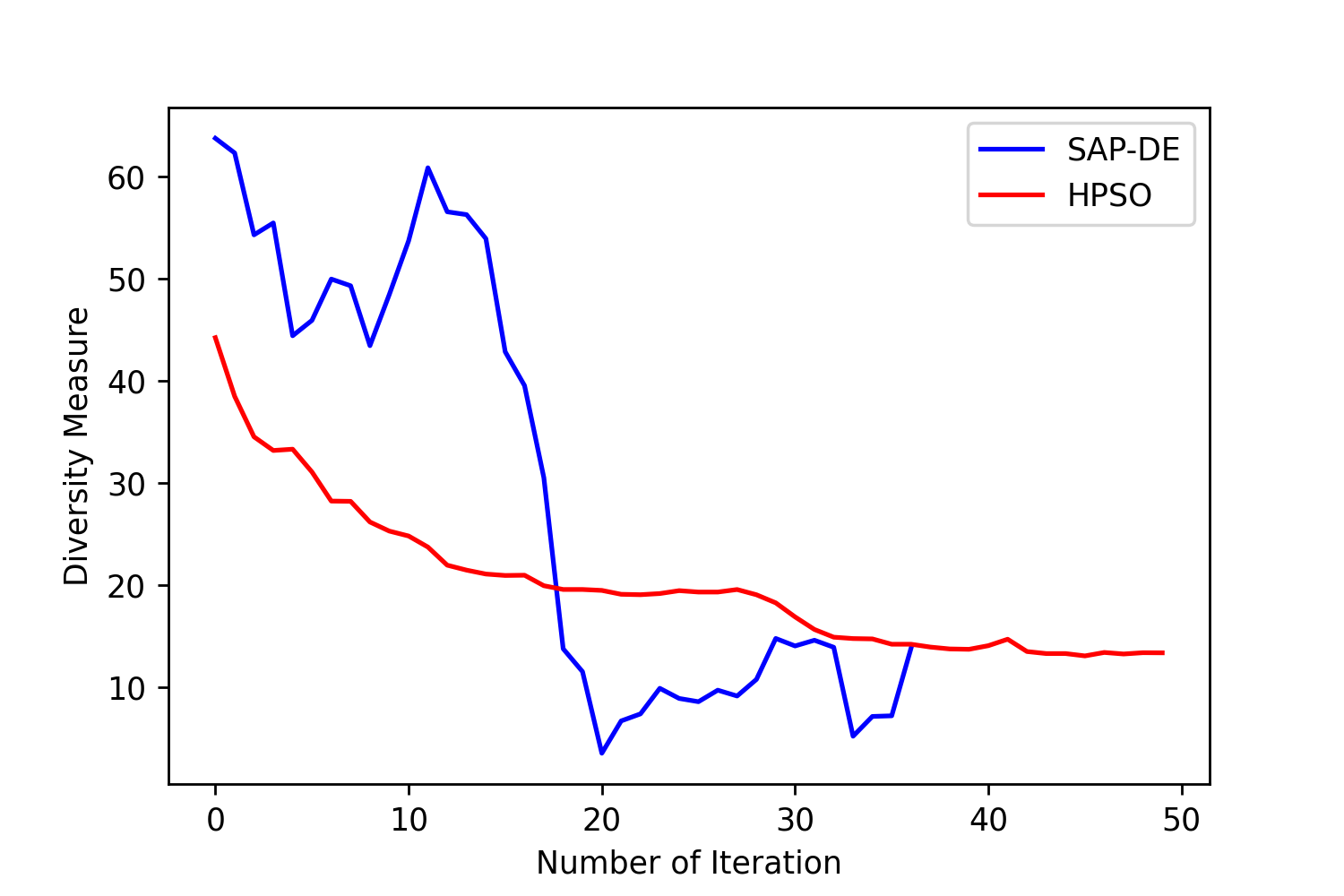}
		\caption{Beach}
	\end{subfigure}
	\begin{subfigure}[b]{0.3\linewidth}
		\includegraphics[width=\linewidth] {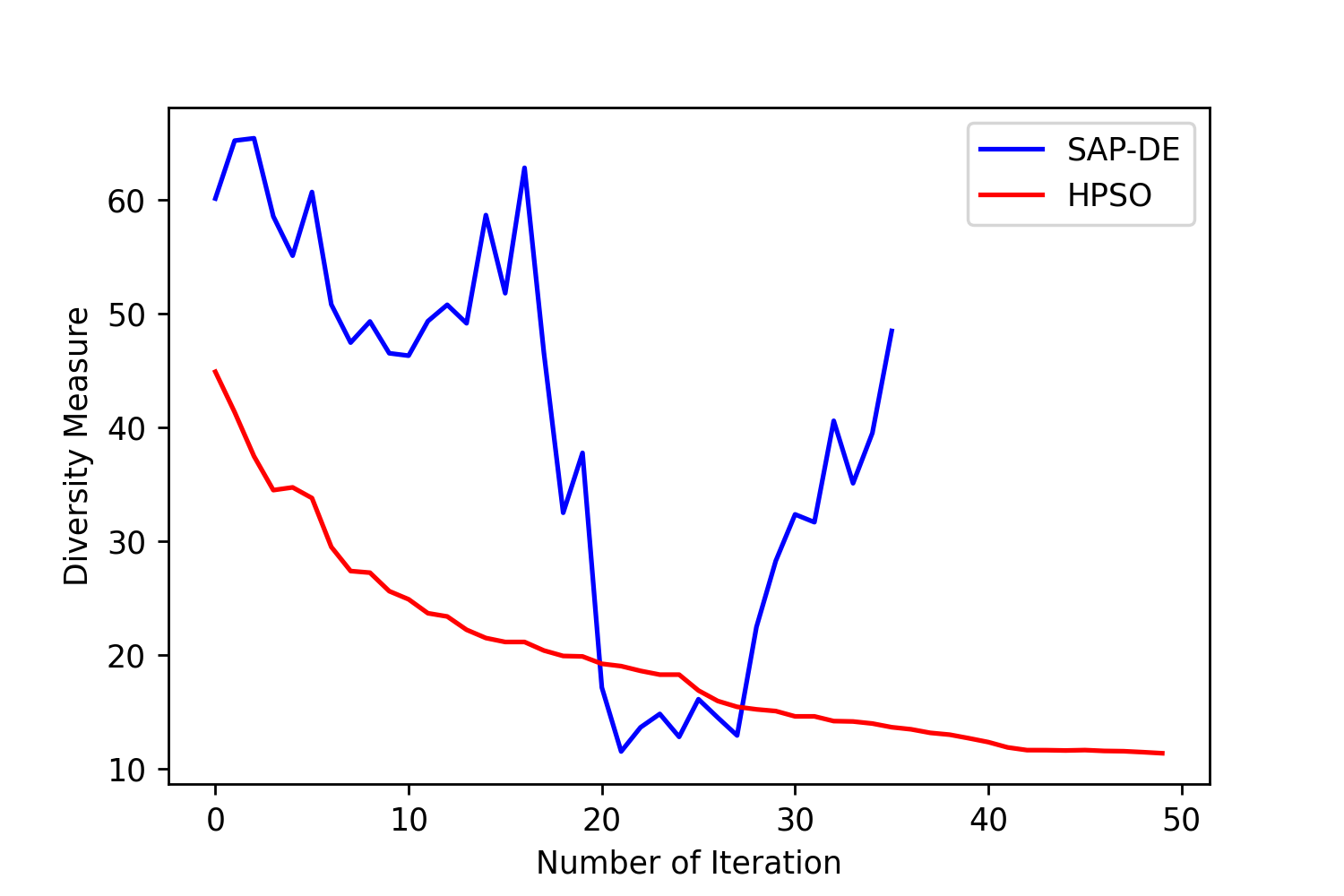}
		\caption{Cathedrals beach}
	\end{subfigure} 
	\begin{subfigure}[b]{0.3\linewidth}
		\includegraphics[width=\linewidth] {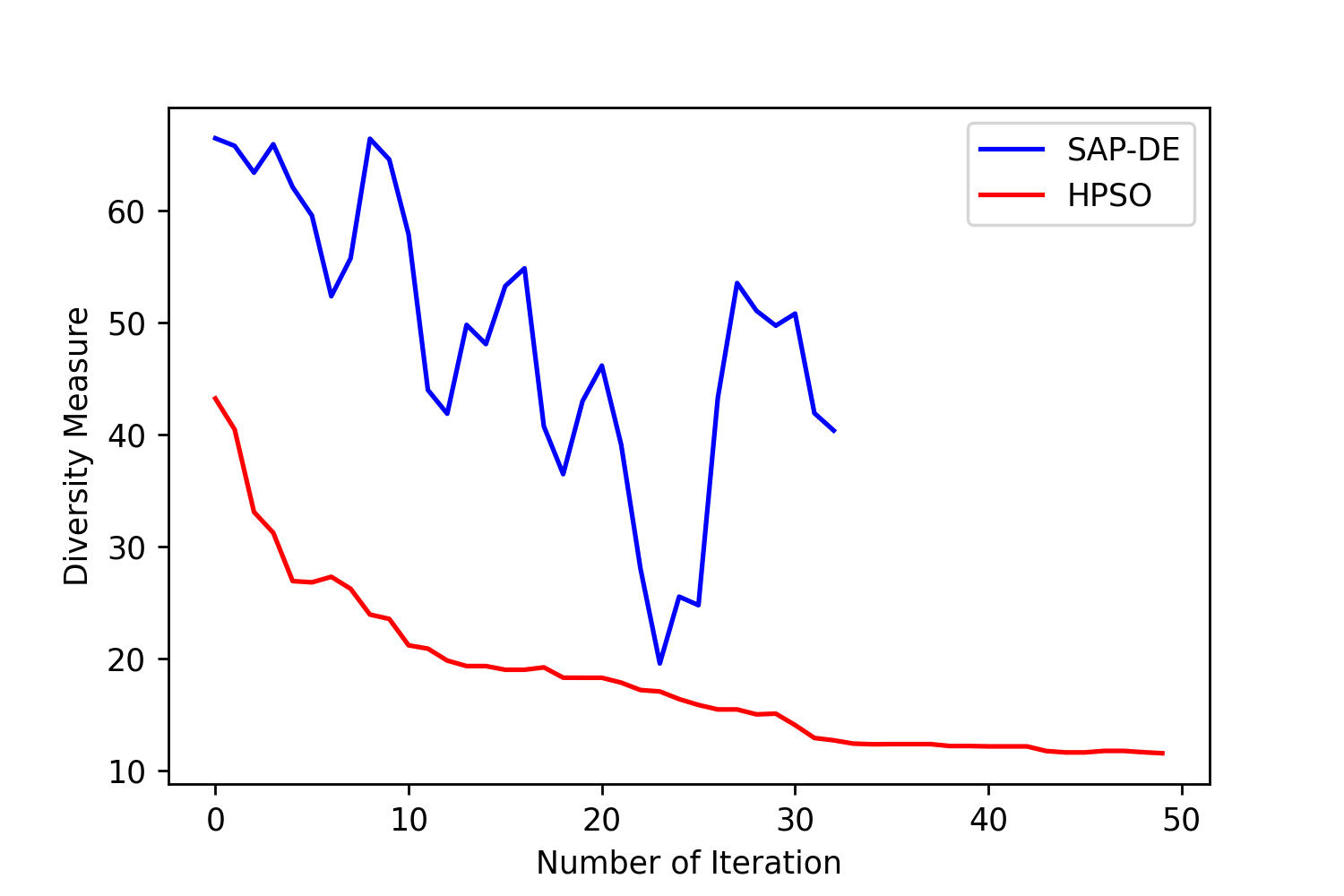}
		\caption{dessert}
	\end{subfigure}
	\begin{subfigure}[b]{0.3\linewidth}
		\includegraphics[width=\linewidth] {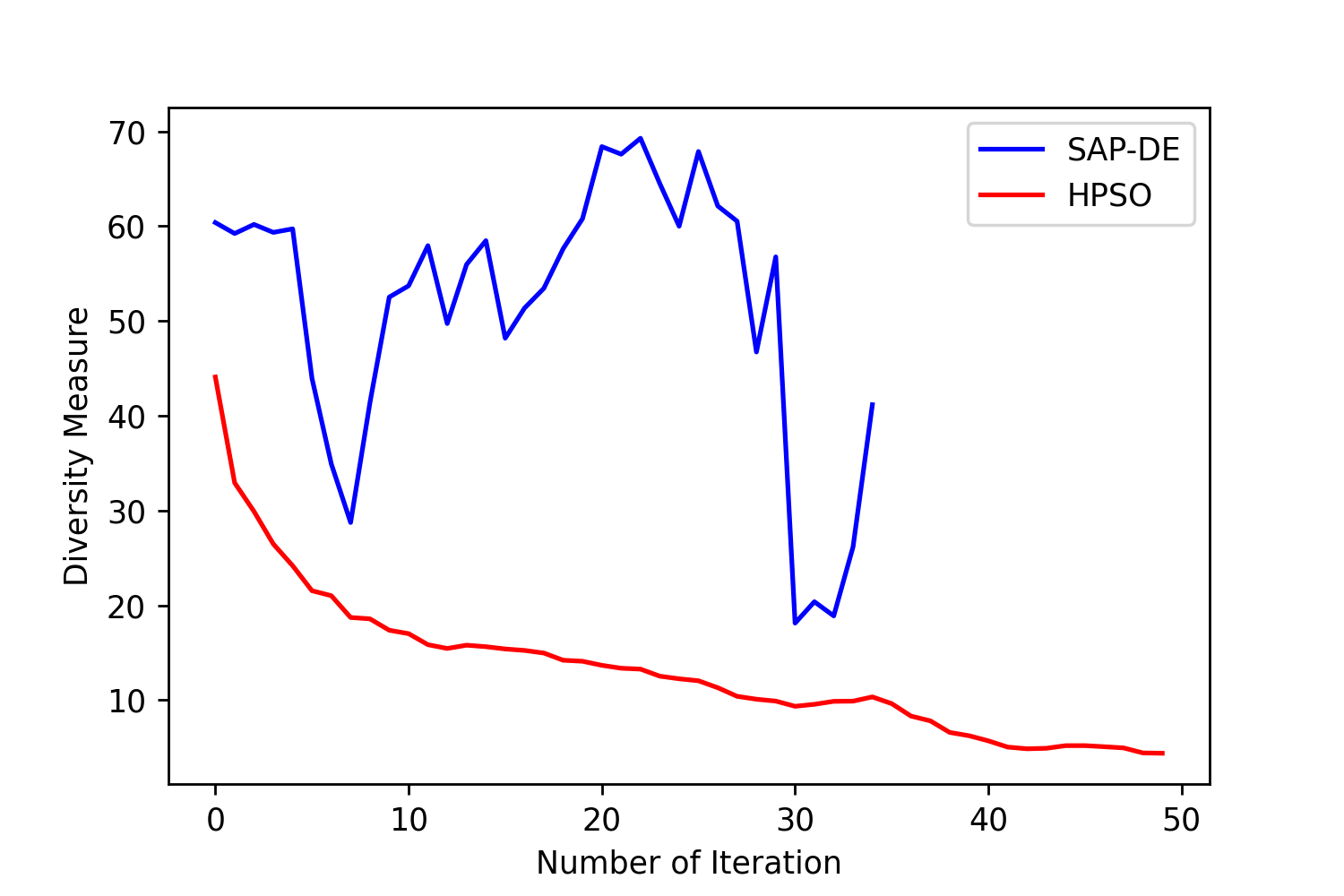}
		\caption{headbands}
	\end{subfigure}
	\begin{subfigure}[b]{0.3\linewidth}
		\includegraphics[width=\linewidth] {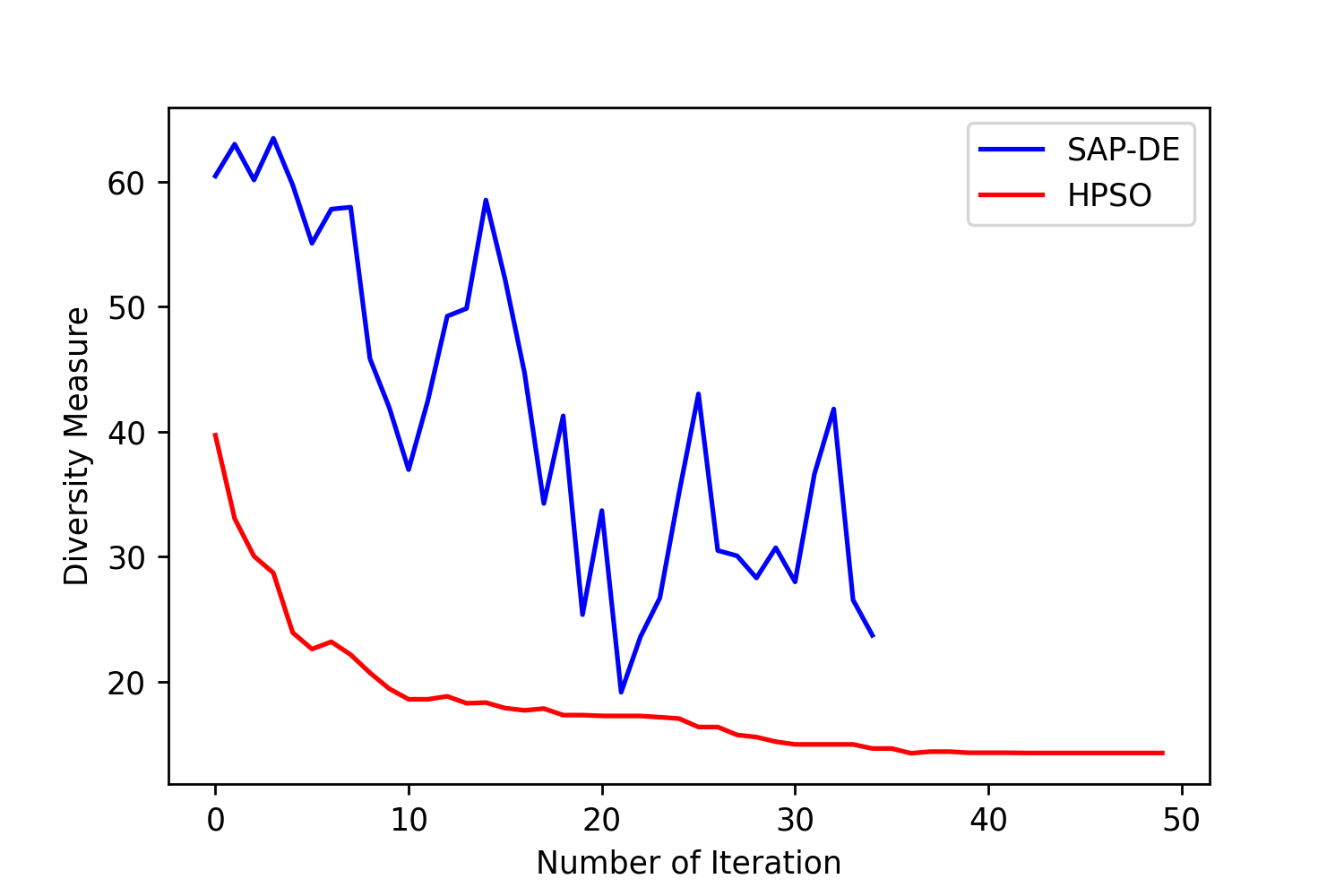}
		\caption{landscape}
	\end{subfigure} 
	
	\caption{Diversity measure.}
	\label{fig:diversity}
\end{figure}

To this end, we selected two algorithms as representatives, HPSO as one of the best-performing algorithms and SAP-DE as one of the worst-performing algorithms. Figure~\ref{fig:diversity} shows the population diversity during the optimisation process for two representative algorithms for all images. Since the stopping criterion in this paper is defined as the number of function evaluations, and SAP-DE in each iteration calculates more objective functions than the population size, the number of iterations for SAP-DE is lower than HPSO. It can be seen that the diversity of HPSO gradually decreases, meaning that in the early stages of the optimisation process, exploration is high, while over the iterations, exploitation is enhanced and exploration is degraded. In all cases, SAP-DE is significantly fluctuating. Even in some cases such as \textit{Snowman}, the trend is upward. In some other cases such as \textit{Cathedrals beach}, there is a downward trend followed by an upward trend.    

The performance of an optimisation algorithm is significantly influenced by its exploration and exploitation capabilities. The two competing goals should be ideally balanced by a competent optimization technique~\cite{PSO_Diversity01}, ~\cite{PSO_Diversity02}. When the exploitation process predominates, the population quickly loses its diversity, and the algorithm quickly converges to a local optimal solution.
On the other hand, if the exploration phase dominates, the algorithm spends a lot of time exploring un-necessary regions of the search space. 
\cite{Exploration_Exploitation} introduced two criteria to measure exploration and exploitation. To this end, first, the center of population is calculated based on a median as
\begin{equation}
Div_{j}=\frac{1}{NP} \sum_{i=1}^{NP} median(x^{i})-x_{i}^{j}
\end{equation}

\begin{equation}
Div=\frac{1}{D} \sum_{j=1}^{D} Div_{j}
\end{equation}

where $median(x^{i})$ is median of dimension $j$ in the entire population, $x_{i}^{j}$ is the dimension $j$ of $i$ candidate solution, $NP$ is the population size, and $D$ is the dimensionality of the problem.

The percentage of exploitation and exploration in one algorithm, for each iteration, can be calculated as

\begin{equation}
XPL=\frac{Div}{Div_{max}} \times 100
\end{equation}
and
\begin{equation}
XPT=\frac{Div-Div_{max}}{Div_{max}} \times 100
\end{equation}

where $Div_{max}$ devotes the maximum diversity in all iterations, $XPL$ and $XPT$ mean exploration and exploitation percentages for an iteration, respectively. 

\begin{figure}[tb!]
	\centering
	\begin{subfigure}[b]{0.30\linewidth}
		\includegraphics[width=\linewidth] {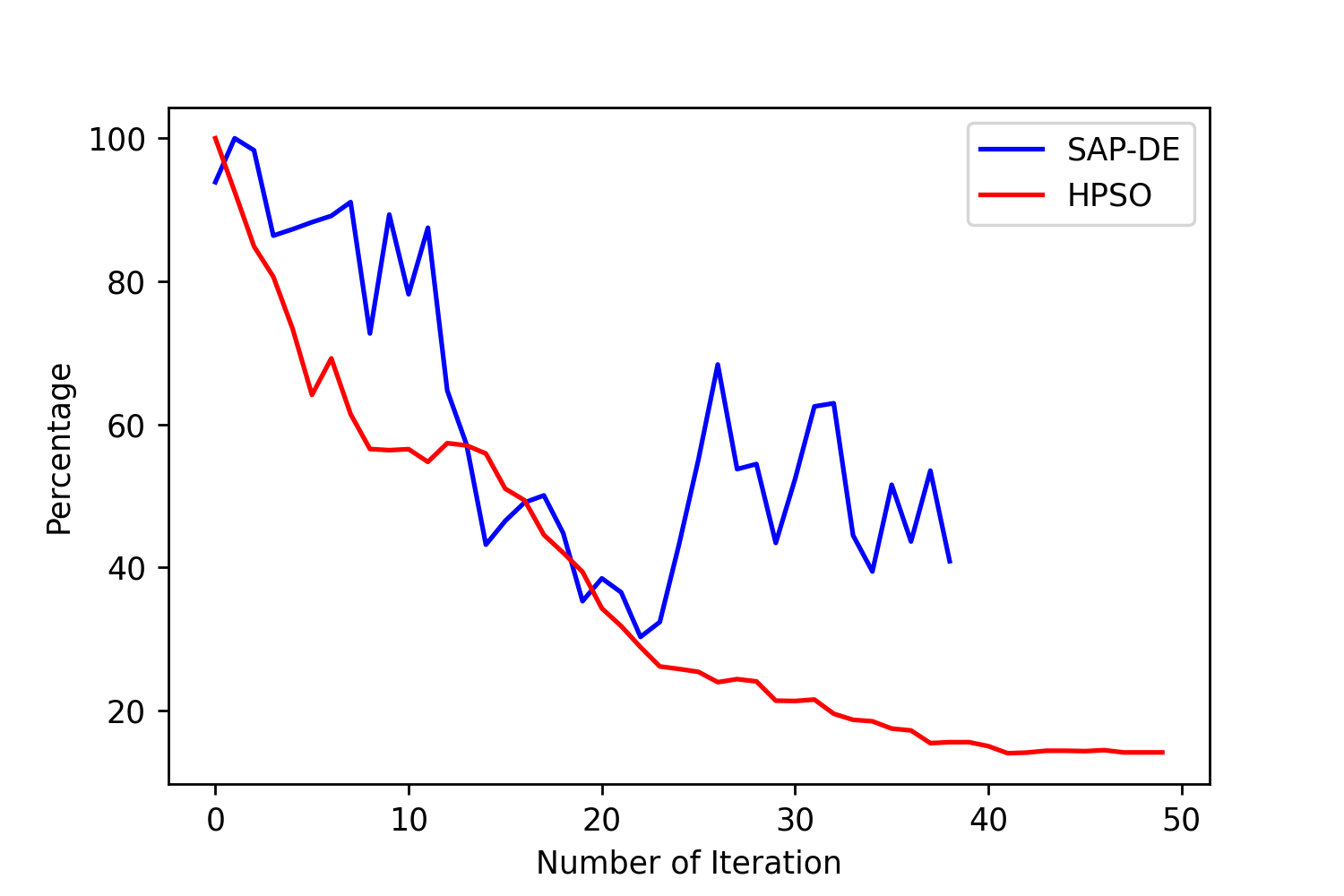}
		\caption{Airplane}
	\end{subfigure} 
	\begin{subfigure}[b]{0.30\linewidth}
		\includegraphics[width=\linewidth] {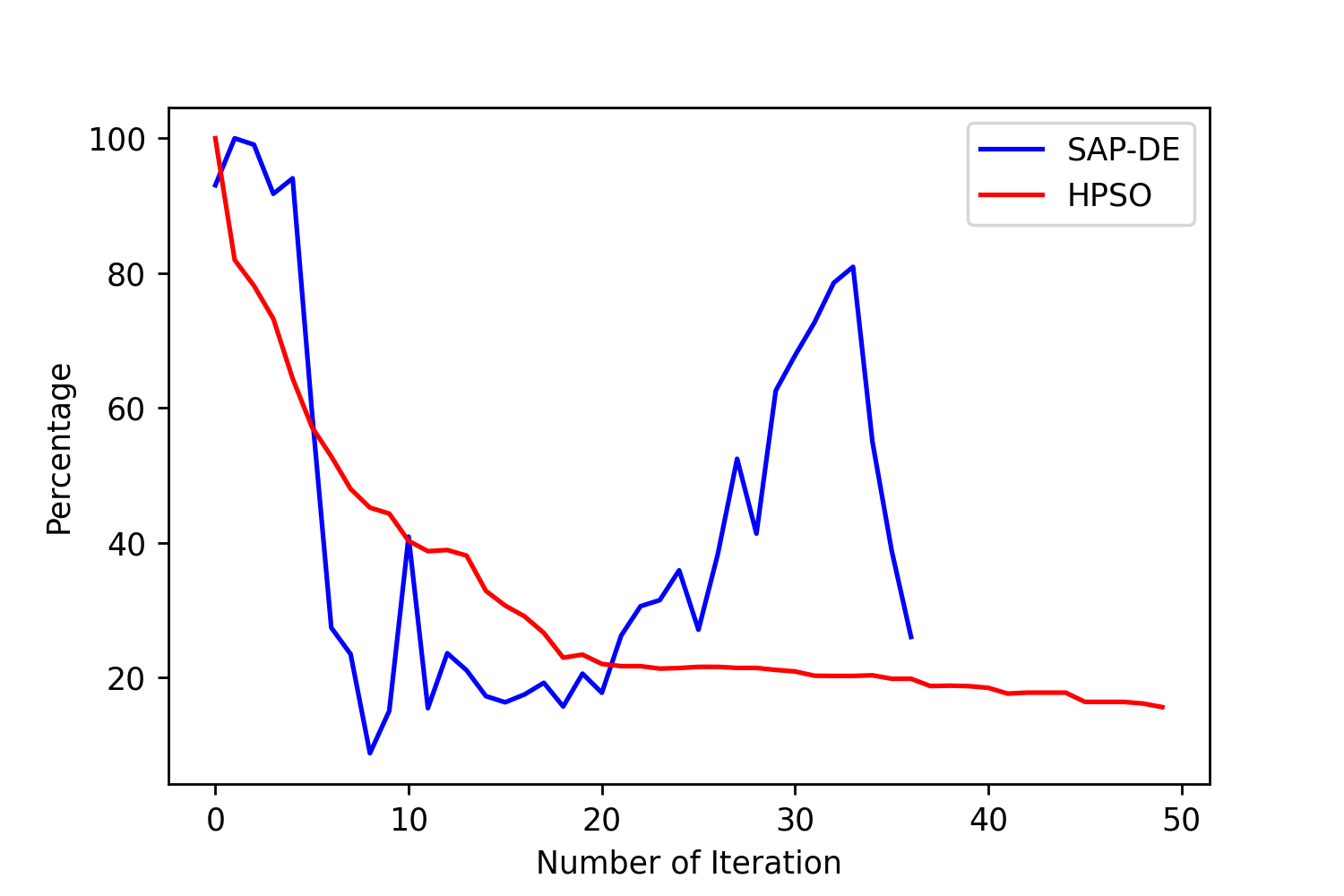}
		\caption{Barbara}
	\end{subfigure}
	\begin{subfigure}[b]{0.30\linewidth}
		\includegraphics[width=\linewidth] {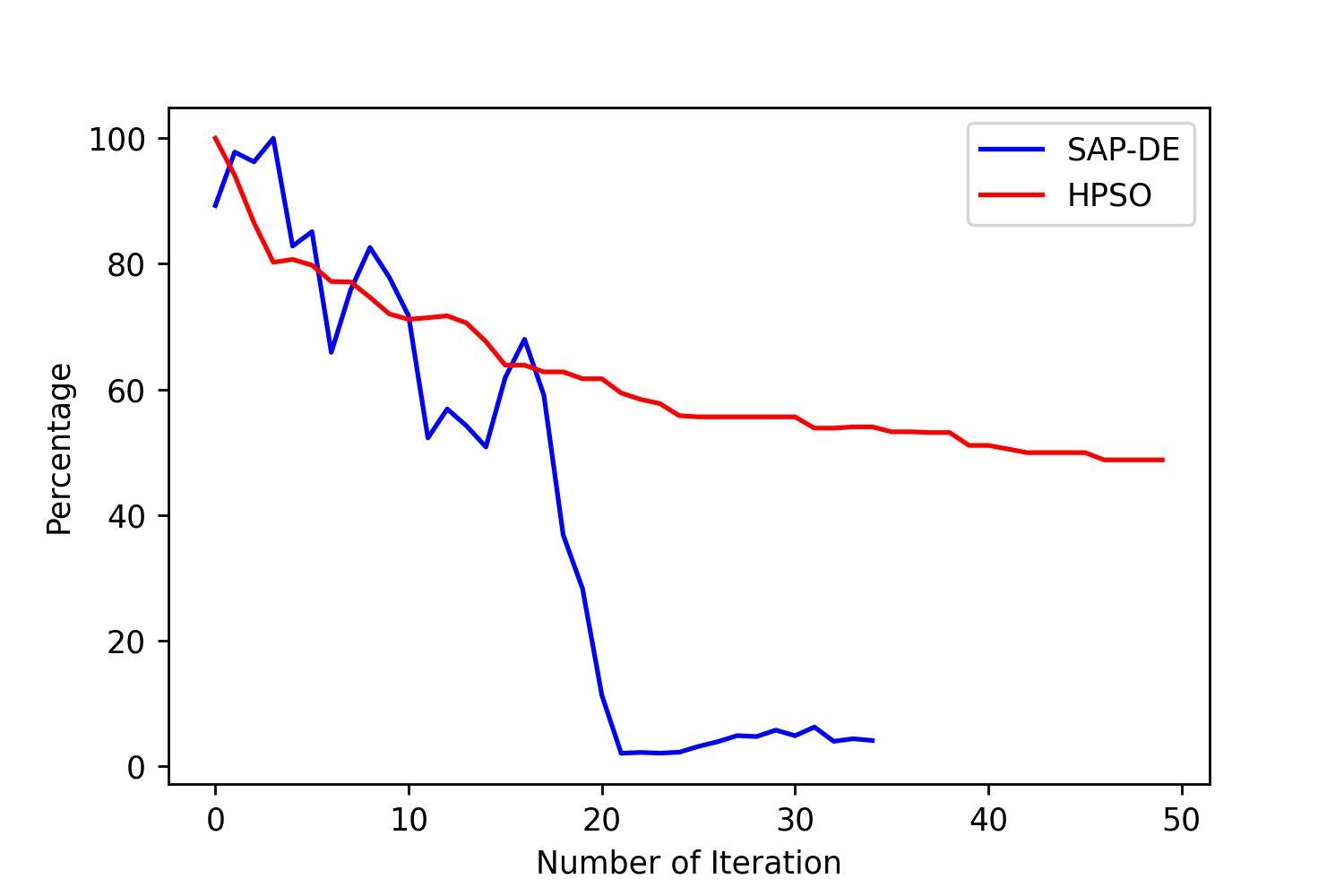}
		\caption{Barbara}
	\end{subfigure}
	\begin{subfigure}[b]{0.30\linewidth}
		\includegraphics[width=\linewidth] {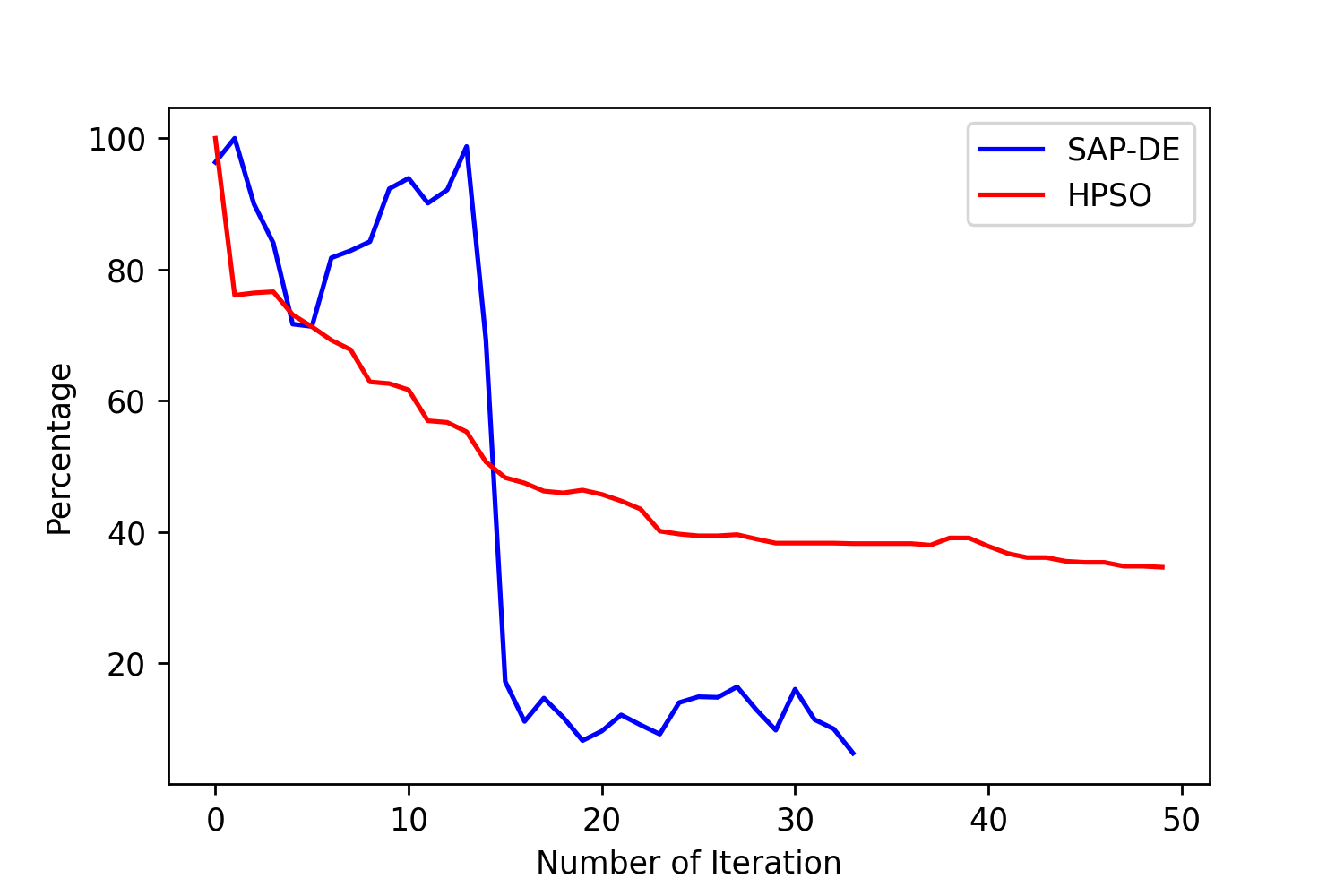}
		\caption{Mandrill}
	\end{subfigure} 
	\begin{subfigure}[b]{0.3\linewidth}
		\includegraphics[width=\linewidth] {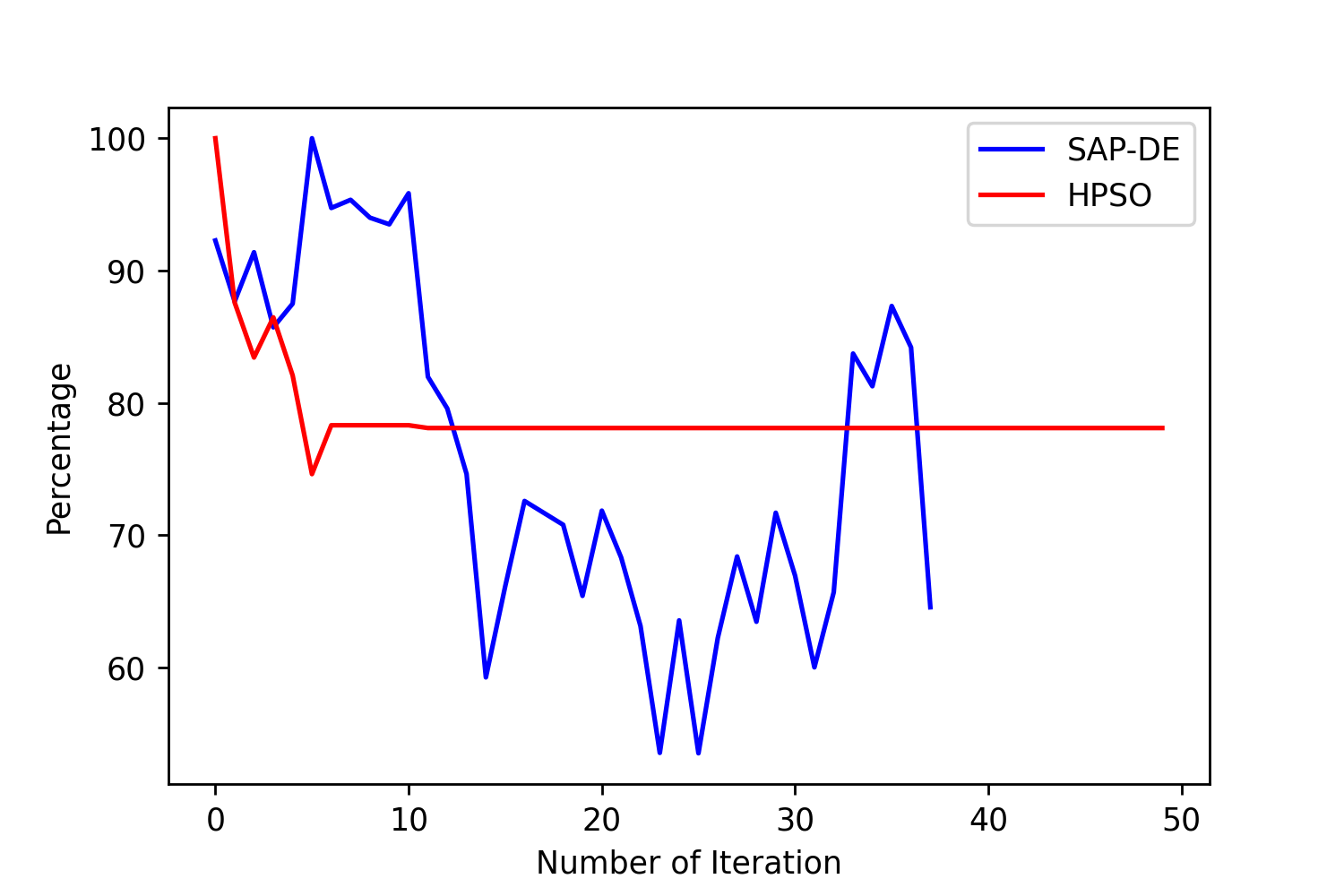}
		\caption{Peppers}
	\end{subfigure}
	\begin{subfigure}[b]{0.3\linewidth}
		\includegraphics[width=\linewidth] {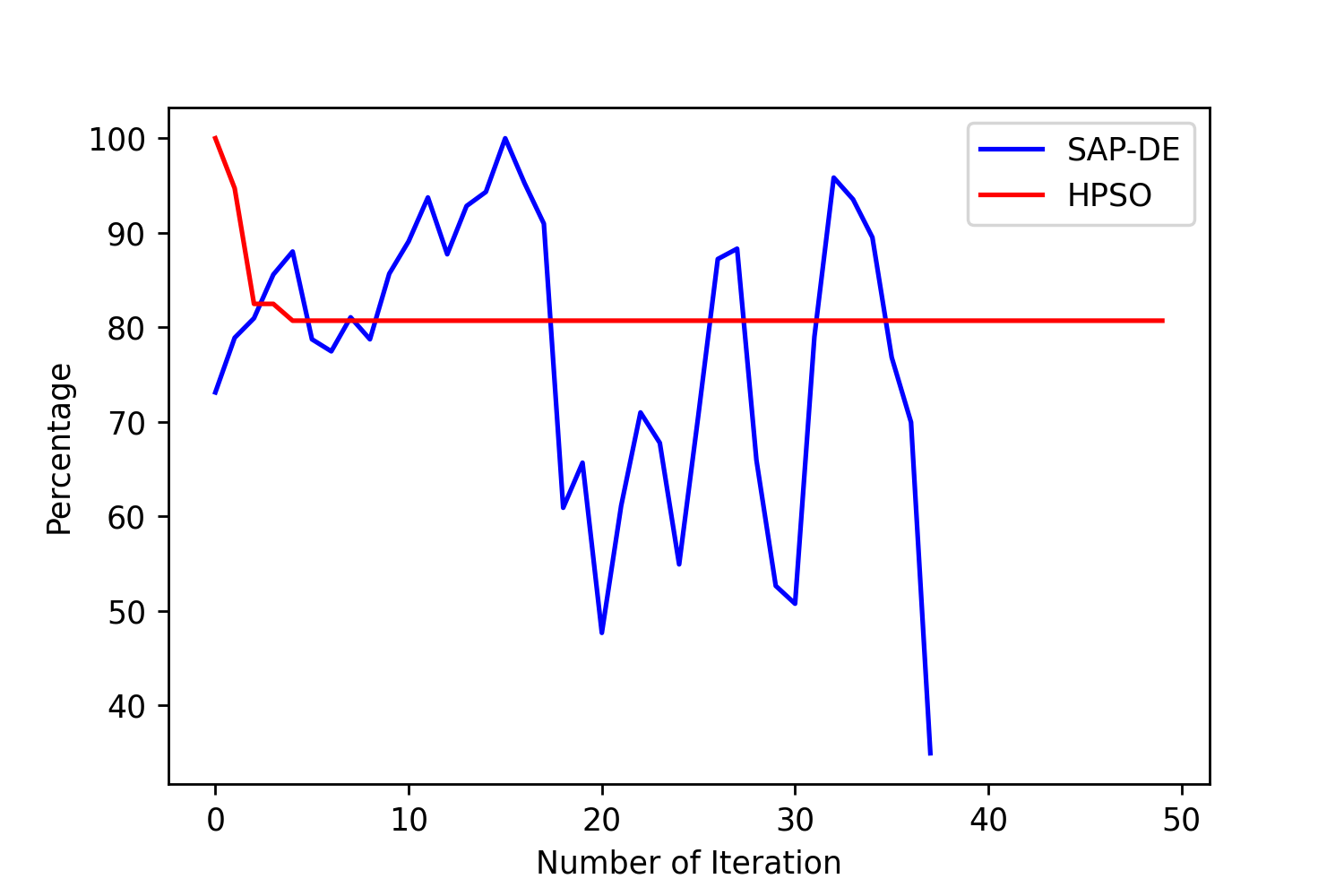}
		\caption{Sailboat}
	\end{subfigure} 
	\begin{subfigure}[b]{0.3\linewidth}
		\includegraphics[width=\linewidth] {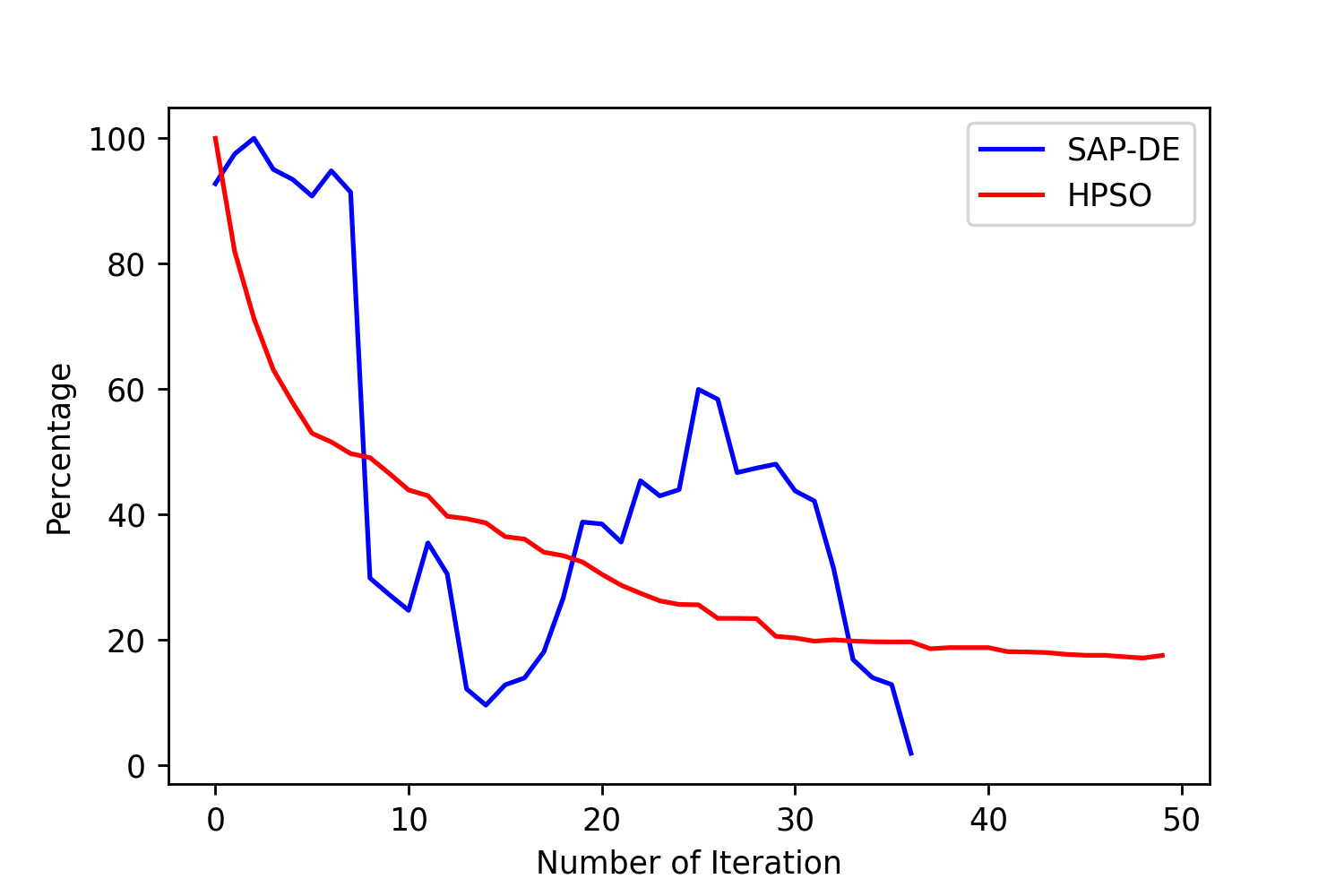}
		\caption{Snowman} 
	\end{subfigure} 	
	\begin{subfigure}[b]{0.3\linewidth}
		\includegraphics[width=\linewidth] {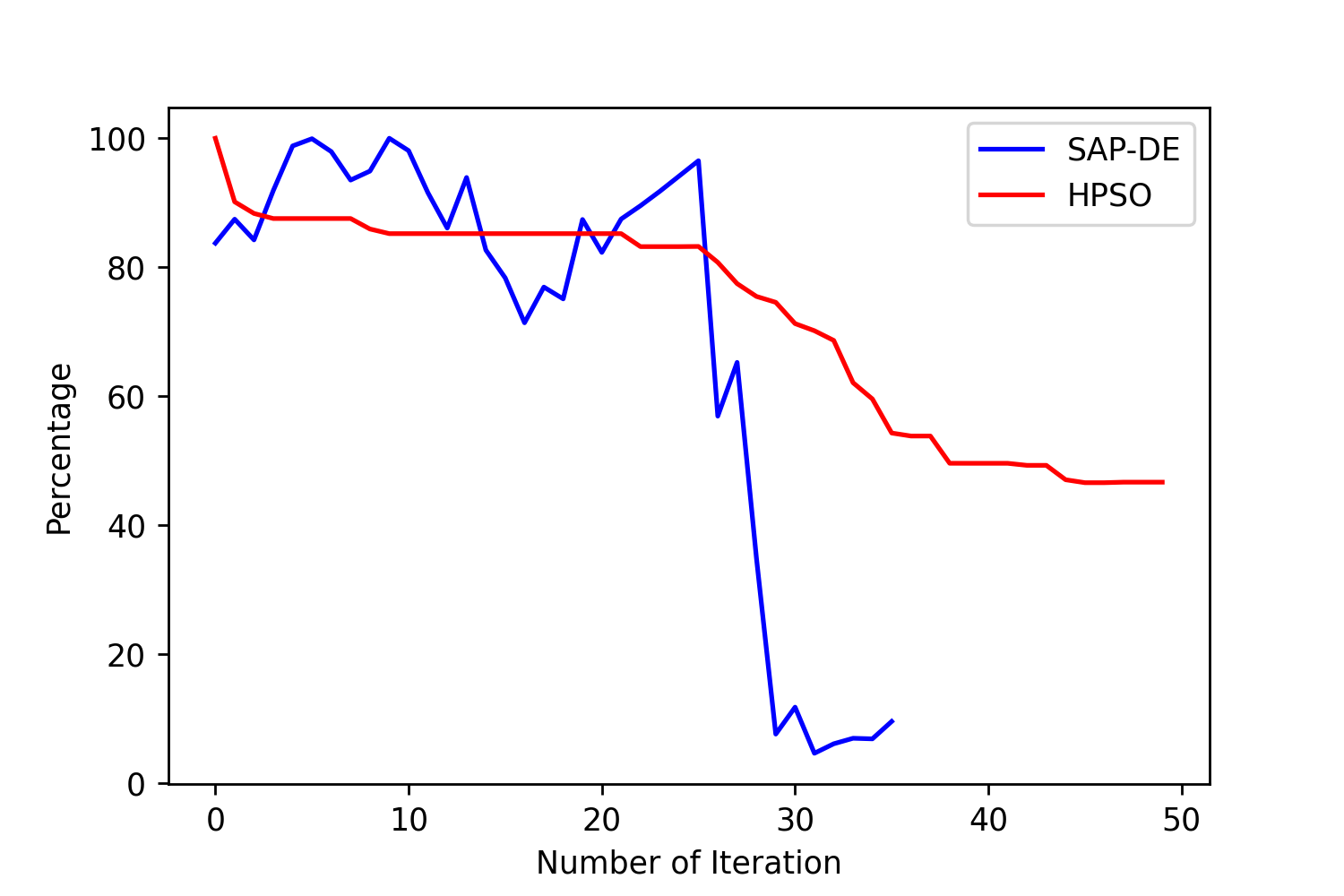}
		\caption{Tiffany}
	\end{subfigure} 
	\begin{subfigure}[b]{0.3\linewidth}
		\includegraphics[width=\linewidth] {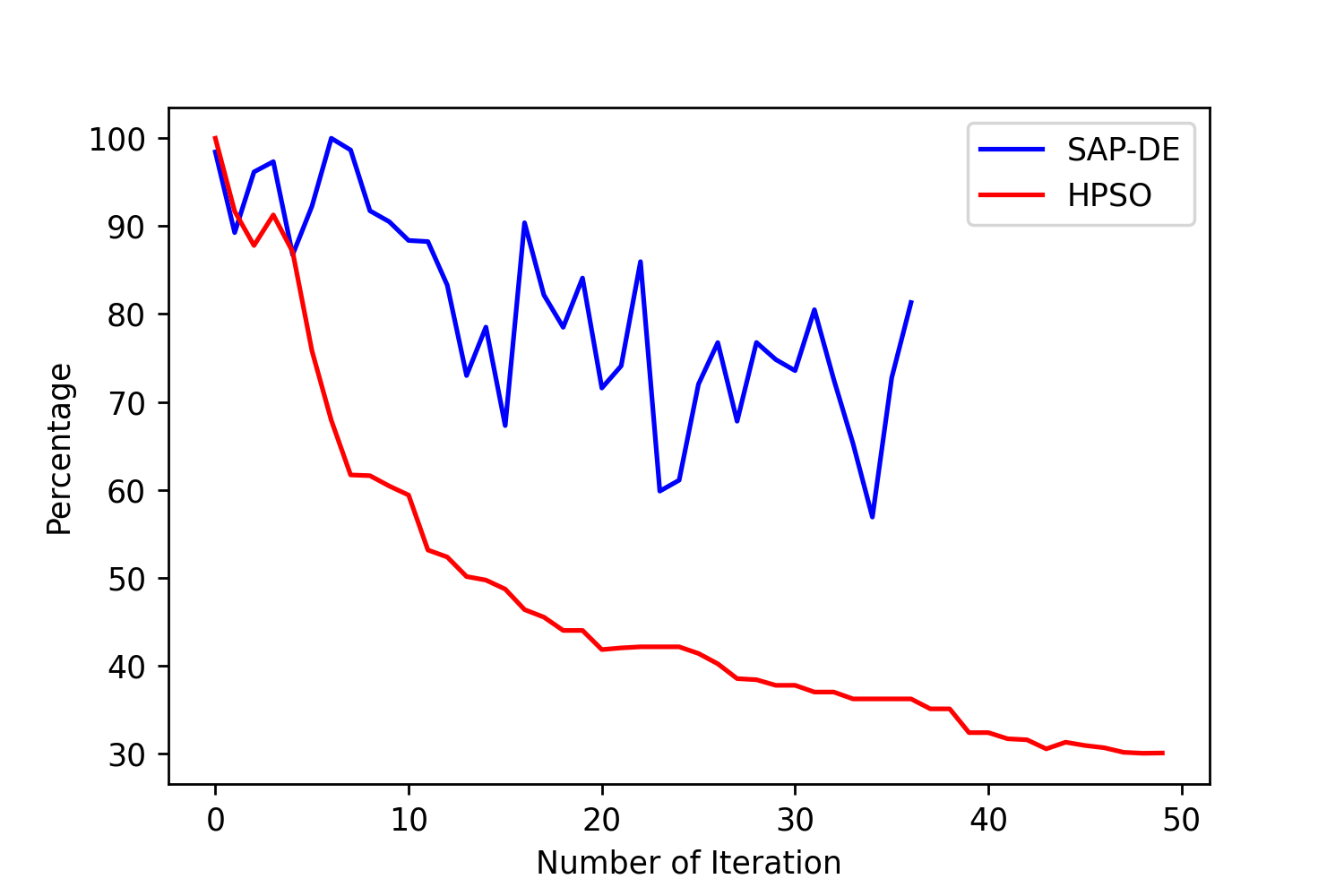}
		\caption{Beach}
	\end{subfigure}
	\begin{subfigure}[b]{0.3\linewidth}
		\includegraphics[width=\linewidth] {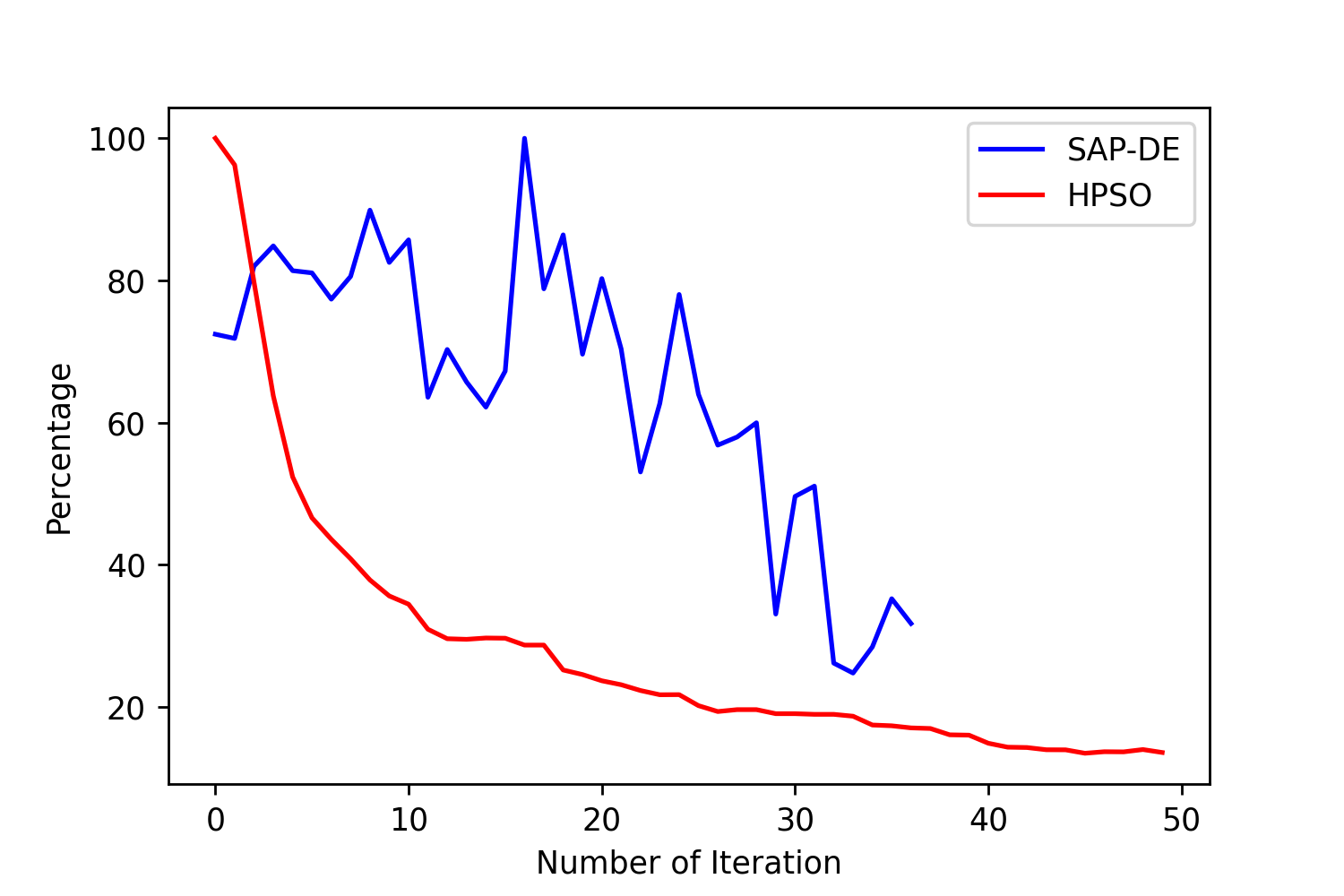}
		\caption{Cathedrals beach}
	\end{subfigure} 
	\begin{subfigure}[b]{0.3\linewidth}
		\includegraphics[width=\linewidth] {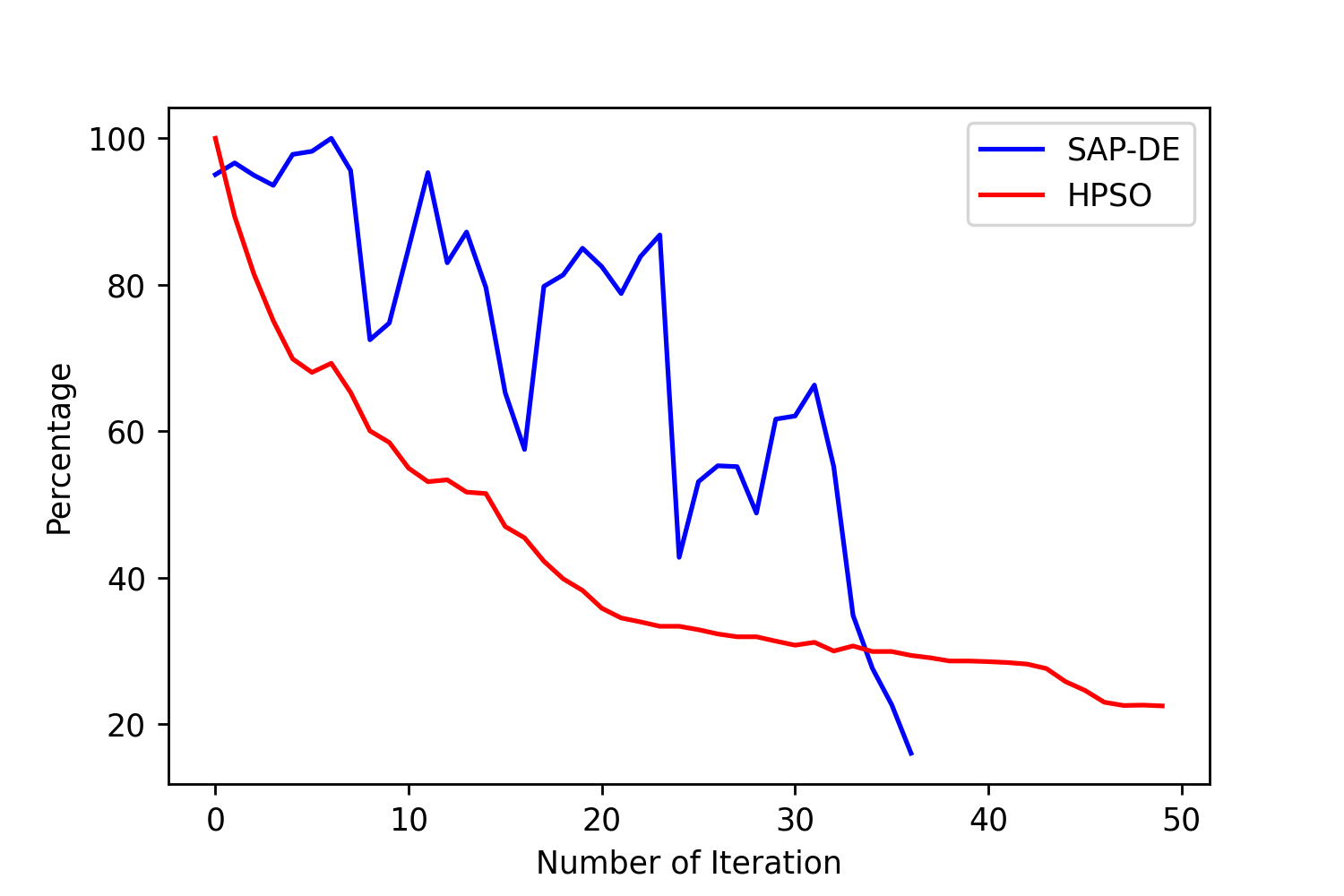}
		\caption{dessert}
	\end{subfigure}
	\begin{subfigure}[b]{0.3\linewidth}
		\includegraphics[width=\linewidth] {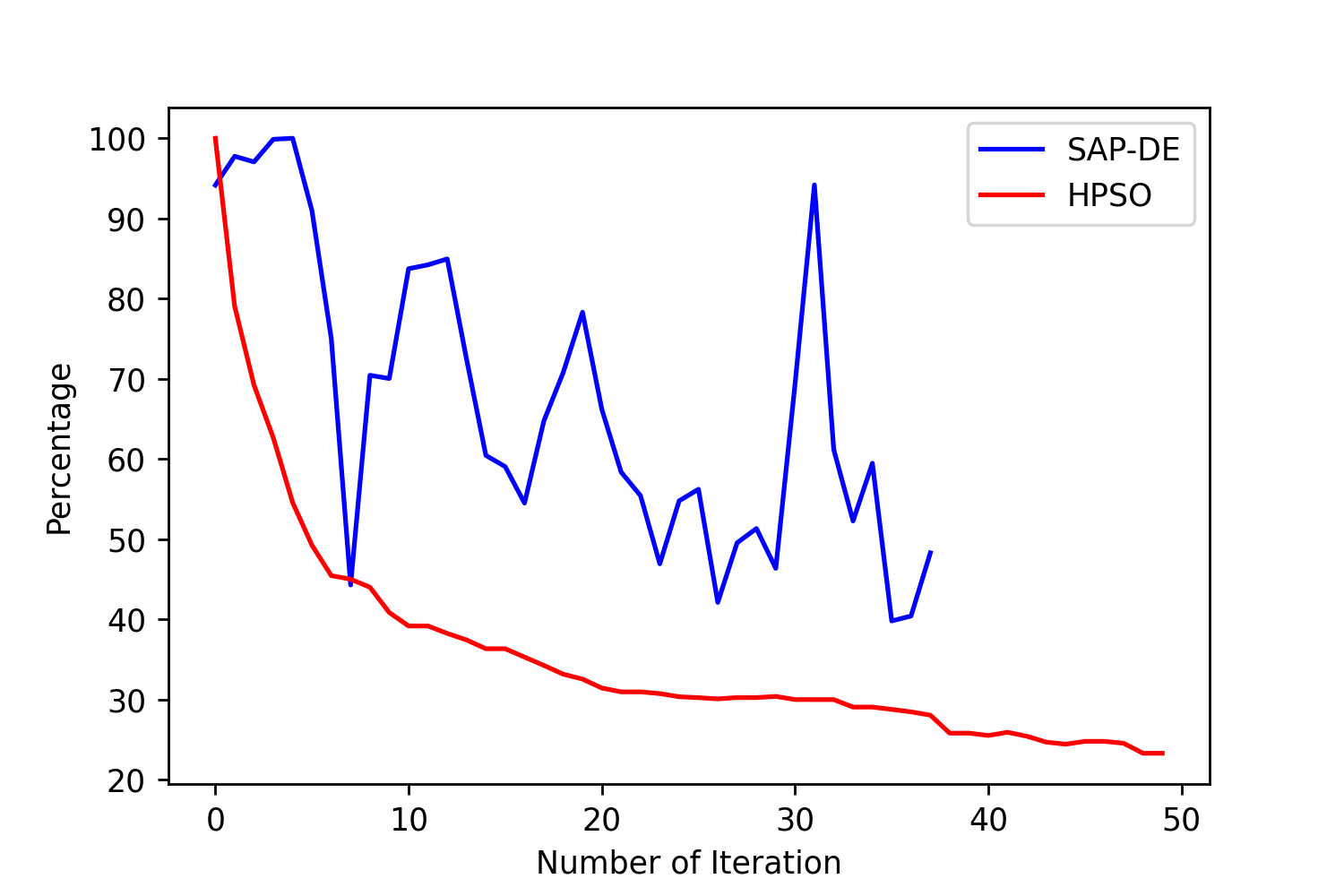}
		\caption{headbands}
	\end{subfigure}
	\begin{subfigure}[b]{0.3\linewidth}
		\includegraphics[width=\linewidth] {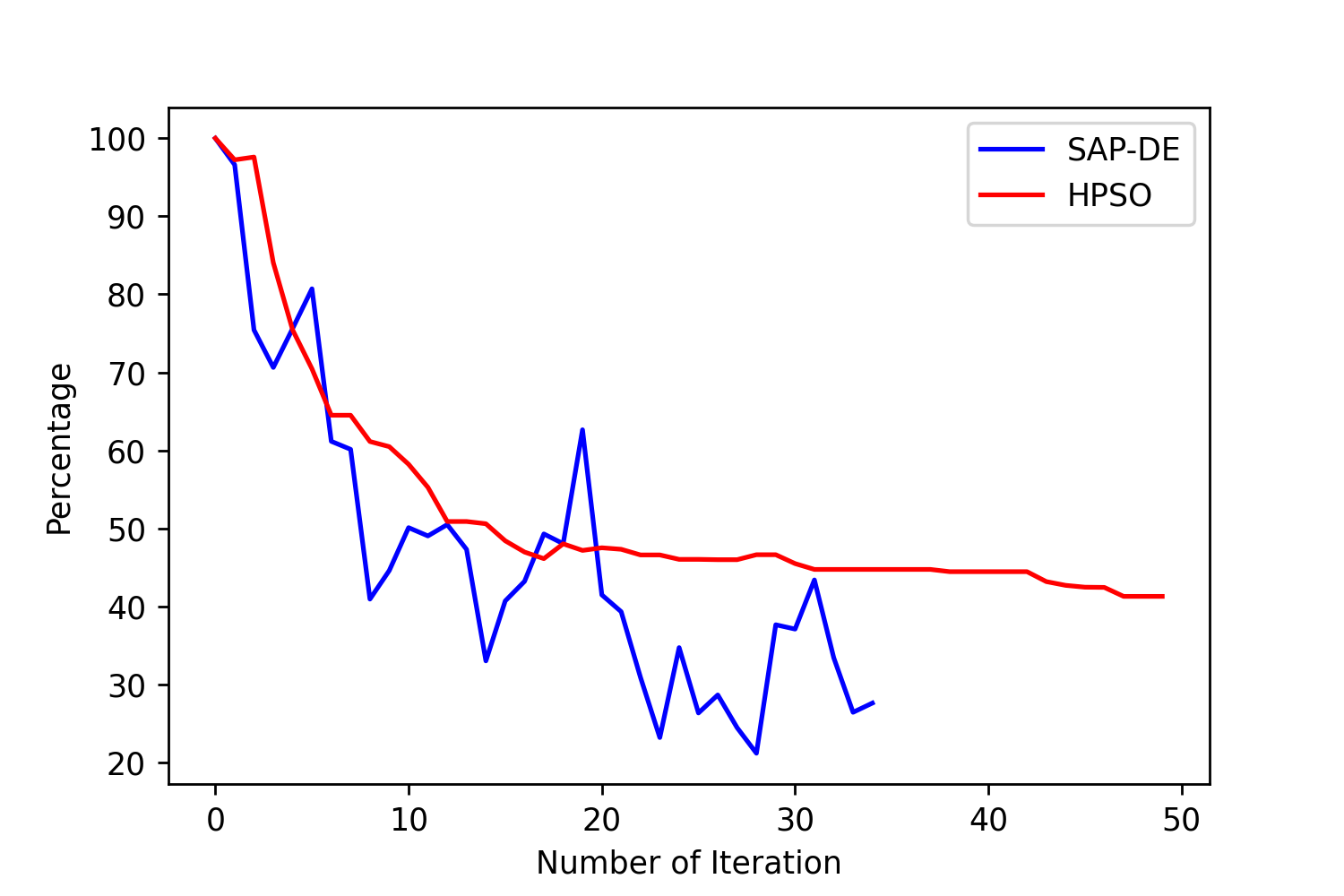}
		\caption{landscape}
	\end{subfigure} 
	
	\caption{Exploration measure.}
	\label{fig:exploration}
\end{figure}

Figure~\ref{fig:exploration} shows the exploration measure for all images and two representatives. It clearly indicates that HPSO gradually decreases the exploration over the course of iteration, while there are drastic fluctuations in the SAP-DE algorithm. Since the exploitation criterion is the opposite, we do not include them in the paper. Therefore, we can say that SAP-DE can not provide enough and regular exploration and exploitation over time, and as a result, its performance is degraded.  

\section{Conclusion}
\label{sec:Conc}
The JPEG standard is one of the most widely employed algorithms in image processing. The quantisation table (QT) influences the image properties such as file size and image quality. Several studies suggest that population-based metaheuristic (PBMH) algorithms can be used to find the right values for QT(s). However, our study shows that these algorithms have three main problems. First, they do not take into account the user opinion, second, the current works can not give an adequate cover of the entire search space, and third, the quality factor (CF) in PBMH-based JPEG image compression algorithms should be determined in advance. To tackle these problems, we re-formulated the population-based JPEG image compression, so that both representation and objective function are changed. By changing the objective function, we incorporated the user opinion on image file size. In other words, file size can be controlled in advanced by a user. In addition, our new representation can solve the problem of lack of a comprehensive coverage. Another benefit of our new representation is that the quality factor can be selected automatically. Since both objective functions and representation are independent of the PBMH, any type of PBMH can be employed to this end. As the fourth contribution, this paper benchmarks 22 PBMHs, both state-of-the-art and newly-introduced algorithms, for the new formulation of JPEG image compression. 

Despite the effectiveness of the proposed strategy, this work can be extended in the future with the following points:

\begin{itemize}
	\item This paper shows that the search strategy can have a significant impact on the final performance. Therefore, more effective search strategies can be used in the future.
	\item This paper incorporates the file size as the user opinion, while other opinions also can be added to the proposed approach. 
	\item A multi-objective variant of the proposed approach is under investigation. 
\end{itemize} 

\section{Acknowledgement}
This work was financed by FEDER (Fundo Europeu de Desenvolvimento Regional), from the European Union through CENTRO 2020 (Programa Operacional Regional do Centro), under project CENTRO-01-0247-FEDER-047256 – GreenStamp: Mobile Energy Efficiency Services.

This work was supported by NOVA LINCS (UIDB/04516/2020) with the financial support of FCT-Fundação para a Ciência e a Tecnologia, through national funds.

\bibliographystyle{elsarticle-num-names}
\bibliography{E:/Dropbox/Dropbox/jalal}

\clearpage
\appendix
\section{Appendix}

\setcounter{table}{0}
\renewcommand{\thetable}{A\arabic{table}}

\begin{table}[!htbp]
	\center
	\caption{Parameter settings for all algorithms.}
	\label{tab:parameter}
\begin{tabular}{ccc}
		\hline
		Algorithms                                    & Parameter                     & Value                    \\ \hline
		\multirow{4}{*}{GA}                           & PC                            & 0.95                     \\ \cline{2-3} 
		& PM                            & 0.05                     \\ \cline{2-3} 
		& selection                     & Tournament               \\ \cline{2-3} 
		& crossover                     & Multi points             \\ \hline
		\multirow{7}{*}{DE, SADE, SAP-DE, JADE}      & weighting factor              & 0.8                      \\ \cline{2-3} 
		& crossover rate                & 0.9                      \\ \cline{2-3} 
		& Strategy                      & DE/current-to-rand/1/bin \\
\hline
	\end{tabular}
\end{table}

\begin{table}[!htbp]
	\centering
	\caption{Mean objective functions for the base algorithms.}
	\label{tab:obj_base}
	\begin{tabular}{l|l|c|cccccc}
		\hline
		Images                            & File Size              &      & GA   & DE   & MA   & PSO  & ES   & ABC  \\ \hline
		\multirow{4}{*}{Airplane}         & \multirow{2}{*}{10000} & Avg. & 0.90 & 1.26 & 1.07 & 1.14 & 0.93 & 0.92 \\
		&                        & R    & 1    & 6    & 4    & 5    & 3    & 2    \\ \cline{2-9}
		& \multirow{2}{*}{50000} & Avg. & 0.81 & 0.87 & 0.88 & 0.86 & 0.90 & 0.83 \\
		&                        & R    & 1    & 4    & 5    & 3    & 6    & 2    \\ \hline
		\multirow{4}{*}{Barbara}          & \multirow{2}{*}{10000} & Avg. & 0.52 & 0.66 & 0.58 & 0.67 & 0.54 & 0.54 \\
		&                        & R    & 1    & 5    & 4    & 6    & 2.5  & 2.5  \\ \cline{2-9}
		& \multirow{2}{*}{50000} & Avg. & 0.45 & 0.54 & 0.53 & 0.54 & 0.54 & 0.46 \\
\hline
	\end{tabular}
\end{table}

\begin{table}[!htbp]
	\centering
	\caption{Closeness measure for the base algorithms.}
	\label{tab:diff_base}
	\begin{tabular}{l|l|c|cccccc}
		\hline
		Images                            & File Size              &      & GA     & DE      & MA      & PSO     & ES      & ABC    \\ \hline
		\multirow{4}{*}{Airplane}         & \multirow{2}{*}{10000} & Avg. & 29.63  & 3635.60 & 1703.70 & 2432.77 & 353.80  & 192.40 \\
		&                        & R    & 1      & 6       & 4       & 5       & 3       & 2      \\ \cline{2-9}
		& \multirow{2}{*}{50000} & Avg. & 145.30 & 1811.70 & 1460.13 & 1186.23 & 4.13    & 422.10 \\
		&                        & R    & 2      & 6       & 5       & 4       & 1       & 3      \\ \hline
		\multirow{4}{*}{Barbara}          & \multirow{2}{*}{10000} & Avg. & 23.43  & 1277.20 & 391.77  & 1339.53 & 41.00   & 163.47 \\
		&                        & R    & 1      & 5       & 4       & 6       & 2       & 3      \\ \cline{2-9}
		& \multirow{2}{*}{50000} & Avg. & 40.03  & 3924.27 & 2993.63 & 3968.90 & 42.10   & 359.37 \\
\hline   
	\end{tabular}
\end{table}

\begin{table}[!htbp]
	\centering
	\caption{CF measure for the base algorithms.}
	\label{tab:acc_base}
	\begin{tabular}{l|l|c|cccccc}
		\hline
		Images                            & File Size              &      & GA   & DE   & MA   & PSO  & ES   & ABC  \\ \hline
		\multirow{4}{*}{Airplane}         & \multirow{2}{*}{10000} & Avg. & 1    & 0.03 & 0.43 & 0.33 & 0.93 & 1    \\
		&                        & R    & 1.5  & 6    & 4    & 5    & 3    & 1.5  \\ \cline{2-9}
		& \multirow{2}{*}{50000} & Avg. & 0.97 & 0.3  & 0.4  & 0.57 & 1    & 0.87 \\
		&                        & R    & 2    & 6    & 5    & 4    & 1    & 3    \\ \hline
		\multirow{4}{*}{Barbara}          & \multirow{2}{*}{10000} & Avg. & 1    & 0.57 & 0.9  & 0.43 & 0.97 & 1    \\
		&                        & R    & 1.5  & 5    & 4    & 6    & 3    & 1.5  \\ \cline{2-9}
		& \multirow{2}{*}{50000} & Avg. & 1    & 0.17 & 0.2  & 0.23 & 0.97 & 0.97 \\
\hline
	\end{tabular}
\end{table}

\begin{table}[!htbp]
	\centering
	\caption{The objective function results for the advanced algorithms.}
	\label{tab:obj_adv}
	\begin{tabular}{l|c|c|cccccccc}
		\hline
		Images                            & File Size              &      & LevyES & SADE & SAP-DE & JADE & CPSO & CLPSO & HPSO & PPSO \\ \hline
		\multirow{4}{*}{Airplane}         & \multirow{2}{*}{10000} & Avg. & 0.93   & 0.95 & 1.08    & 0.95 & 0.91 & 0.89  & 0.91 & 0.90 \\
		&                        & R    & 5      & 6.5  & 8       & 6.5  & 3.5  & 1     & 3.5  & 2    \\ \cline{2-11}
		& \multirow{2}{*}{50000} & Avg. & 0.83   & 0.82 & 1.10    & 0.81 & 0.82 & 0.81  & 0.80 & 0.82 \\
		&                        & R    & 7      & 5    & 8       & 2.5  & 5    & 2.5   & 1    & 5    \\ \hline
		\multirow{4}{*}{Barbara}          & \multirow{2}{*}{10000} & Avg. & 0.52   & 0.54 & 0.60    & 0.53 & 0.53 & 0.51  & 0.52 & 0.53 \\
		&                        & R    & 2.5    & 7    & 8       & 5    & 5    & 1     & 2.5  & 5    \\ \cline{2-11}
		& \multirow{2}{*}{50000} & Avg. & 0.46   & 0.45 & 0.57    & 0.45 & 0.46 & 0.45  & 0.44 & 0.46 \\
\hline
	\end{tabular}
\end{table}

\begin{table}[!htbp]
	\small
	\centering
	\caption{Closeness measure for the advanced algorithms.}
	\label{tab:distance_adv}
	\begin{tabular}{l|c|c|cccccccc}
		\hline
		Images                            & File Size              &      & LevyES  & SADE    & SAP-DE & JADE    & CPSO   & CLPSO   & HPSO   & PPSO    \\ \hline
		\multirow{4}{*}{Airplane}         & \multirow{2}{*}{10000} & Avg. & 359.23  & 487.17  & 1834.13 & 487.93  & 121.97 & 23.60   & 184.33 & 9.50    \\
		&                        & R    & 5       & 6       & 8       & 7       & 3      & 2       & 4      & 1       \\ \cline{2-11}
		& \multirow{2}{*}{50000} & Avg. & 7.43    & 313.93  & 2094.23 & 135.40  & 233.03 & 56.00   & 28.63  & 5.67    \\
		&                        & R    & 2       & 7       & 8       & 5       & 6      & 4       & 3      & 1       \\ \hline
		\multirow{4}{*}{Barbara}          & \multirow{2}{*}{10000} & Avg. & 2.50    & 56.57   & 647.57  & 60.10   & 84.63  & 11.57   & 4.67   & 7.13    \\
		&                        & R    & 1       & 5       & 8       & 6       & 7      & 4       & 2      & 3       \\ \cline{2-11}
		& \multirow{2}{*}{50000} & Avg. & 1.80    & 337.17  & 370.33  & 94.87   & 254.27 & 76.50   & 12.03  & 407.63  \\
\hline
	\end{tabular}
\end{table}

\begin{table}[!htbp]
	\centering
	\caption{CF measure results for the advanced algorithms.}
	\label{tab:acc_advanced}
	\begin{tabular}{l|c|c|cccccccc}
		\hline
		Images                            & File Size              &      & LevyES & SADE & SAP-DE & JADE & CPSO & CLPSO & HPSO & PPSO \\ \hline
		\multirow{4}{*}{Airplane}         & \multirow{2}{*}{10000} & Avg. & 0.93   & 0.87 & 0.33    & 0.90 & 1.00 & 1.00  & 0.97      & 1.00 \\
		&                        & R    & 5      & 7    & 8       & 6    & 2    & 2     & 4         & 2    \\ \cline{2-11}
		& \multirow{2}{*}{50000} & Avg. & 1.00   & 0.97 & 0.37    & 1.00 & 1.00 & 1.00  & 1.00      & 1.00 \\
		&                        & R    & 3.5    & 7    & 8       & 3.5  & 3.5  & 3.5   & 3.5       & 3.5  \\ \hline
		\multirow{4}{*}{Barbara}          & \multirow{2}{*}{10000} & Avg. & 1.00   & 1.00 & 0.80    & 1.00 & 1.00 & 1.00  & 1.00      & 1.00 \\
		&                        & R    & 4      & 4    & 8       & 4    & 4    & 4     & 4         & 4    \\ \cline{2-11}
		& \multirow{2}{*}{50000} & Avg. & 1.00   & 0.93 & 0.93    & 1.00 & 1.00 & 1.00  & 1.00      & 0.90 \\
\hline
	\end{tabular}
\end{table}

\begin{table}[!htbp]
	\centering
	\caption{Objective function results for the metaphor-based algorithms.}
	\label{tab:obj_metaphore}
	\begin{tabular}{l|l|l|cccccccc}
		\hline
		Images                            & File Size              &      & HS   & GWO  & ALO  & DA   & WOA  & SCA  & GBO  & AOA  \\ \hline
		\multirow{4}{*}{Airplane}         & \multirow{2}{*}{10000} & Avg. & 0.93 & 0.94 & 0.92 & 1.17 & 0.91 & 0.87 & 0.98 & 1.03 \\
		&                        & R    & 4    & 5    & 3    & 8    & 2    & 1    & 6    & 7    \\ \cline{2-11}
		& \multirow{2}{*}{50000} & Avg. & 0.81 & 0.81 & 0.81 & 0.85 & 0.81 & 0.87 & 0.83 & 0.88 \\
		&                        & R    & 2.5  & 2.5  & 2.5  & 6    & 2.5  & 7    & 5    & 8    \\ \hline
		\multirow{4}{*}{Barbara}          & \multirow{2}{*}{10000} & Avg. & 0.53 & 0.52 & 0.52 & 0.59 & 0.52 & 0.54 & 0.53 & 0.59 \\
		&                        & R    & 4.5  & 2    & 2    & 7.5  & 2    & 6    & 4.5  & 7.5  \\ \cline{2-11}
		& \multirow{2}{*}{50000} & Avg. & 0.44 & 0.45 & 0.46 & 0.50 & 0.44 & 0.54 & 0.46 & 0.54 \\
\hline
	\end{tabular}
\end{table}

\begin{table}[!htbp]
	\centering
	\caption{Closeness measure for the metaphor-based algorithms.}
	\label{tab:distance_metaphor}
	\begin{tabular}{l|l|l|cccccccc}
		\hline
		Images                            & File Size              &      & HS     & GWO    & ALO     & DA      & WOA    & SCA     & GBO     & AOA     \\ \hline
		\multirow{4}{*}{Airplane}         & \multirow{2}{*}{10000} & Avg. & 352.50 & 528.90 & 272.50  & 2727.00 & 184.33 & 1474.53 & 844.43  & 1336.57 \\
		&                        & R    & 3      & 4      & 2       & 8       & 1      & 7       & 5       & 6       \\ \cline{2-11}
		& \multirow{2}{*}{50000} & Avg. & 10.60  & 68.17  & 278.47  & 972.30  & 7.30   & 1474.53 & 383.20  & 2171.80 \\
		&                        & R    & 2      & 3      & 4       & 6       & 1      & 7       & 5       & 8       \\ \hline
		\multirow{4}{*}{Barbara}          & \multirow{2}{*}{10000} & Avg. & 0.57   & 1.47   & 17.20   & 581.27  & 4.67   & 4113.17 & 36.67   & 537.40  \\
		&                        & R    & 1      & 2      & 4       & 7       & 3      & 8       & 5       & 6       \\ \cline{2-11}
		& \multirow{2}{*}{50000} & Avg. & 10.73  & 150.30 & 770.03  & 2293.80 & 16.80  & 4113.17 & 471.83  & 3190.63 \\
\hline
	\end{tabular}
\end{table}

\begin{table}[!htbp]
	\centering
	\caption{CF measure for the metaphor-based algorithms.}
	\label{tab:acc_metaphor}
	\begin{tabular}{l|l|l|cccccccc}
		\hline
		Images                            & File Size              &      & HS   & GWO  & ALO  & DA   & WOA  & SCA  & GBO  & AOA  \\ \hline
		\multirow{4}{*}{Airplane}         & \multirow{2}{*}{10000} & Avg. & 0.93 & 0.90 & 0.97 & 0.27 & 0.97 & 0.50 & 0.83 & 0.50 \\
		&                        & R    & 3    & 4    & 1.5  & 8    & 1.5  & 6.5  & 5    & 6.5  \\ \cline{2-11}
		& \multirow{2}{*}{50000} & Avg. & 1.00 & 1.00 & 1.00 & 0.67 & 1.00 & 0.50 & 0.93 & 0.20 \\
		&                        & R    & 2.5  & 2.5  & 2.5  & 6    & 2.5  & 7    & 5    & 8    \\  \hline
		\multirow{4}{*}{Barbara}          & \multirow{2}{*}{10000} & Avg. & 1.00 & 1.00 & 1.00 & 0.90 & 1.00 & 0.27 & 1.00 & 0.87 \\
		&                        & R    & 3    & 3    & 3    & 6    & 3    & 8    & 3    & 7    \\ \cline{2-11}
		& \multirow{2}{*}{50000} & Avg. & 1.00 & 1.00 & 0.77 & 0.60 & 1.00 & 0.27 & 0.90 & 0.17 \\
\hline
	\end{tabular}
\end{table}

\end{document}